\documentclass[journal]{IEEEtran}
\pdfoutput=1

\let\labelindent\relax
\usepackage{enumitem}

\usepackage{graphicx}
\usepackage{booktabs} 
\usepackage{color}
\usepackage{ifthen}
\usepackage{epstopdf}
\usepackage[scriptsize]{subfigure} 
\usepackage{cite}
\usepackage{amsmath}
\usepackage{amsfonts}
\usepackage{bm}

\usepackage{relsize}

\usepackage{hyperref}
\hypersetup{bookmarks=true,colorlinks=true,bookmarksnumbered=true,bookmarksopen=true}

\usepackage{verbatim}
\usepackage{chngcntr}

\usepackage{pgfplotstable}

\usepackage{xr}
\newcommand{\trsecref}[1]{\cite[Sec.~\begin{NoHyper}\ref{tr-#1}\end{NoHyper}]{tr:tails-2015}}

\definecolor{grey50}{rgb}{0.5,0.5,0.5}
\newcommand{\sgn}{\operatorname{sgn}}

 \newcommand\copyrighttext{%
 \textcopyright 2016 IEEE. Personal use of this material is permitted.
  Permission from IEEE must be obtained for all other uses, in any current or future
  media, including reprinting/republishing this material for advertising or promotional
  purposes, creating new collective works, for resale or redistribution to servers or
  lists, or reuse of any copyrighted component of this work in other works.
  DOI: \href{https://doi.org/10.1109/TRO.2016.2597316}{10.1109/TRO.2016.2597316}}
\newcommand\copyrightnotice{%
\begin{tikzpicture}[remember picture,overlay]
\node[anchor=south,yshift=5pt] at (current page.south) {\fbox{\parbox{\dimexpr\textwidth-\fboxsep-\fboxrule\relax}{  \footnotesize \copyrighttext}}};
\end{tikzpicture}%
}
 
\begin{document}
\bstctlcite{IEEEexample:BSTcontrol}

\title{Comparative Design, Scaling, and Control of Appendages for Inertial~Reorientation}

\author{Thomas~Libby,~\IEEEmembership{Student~Member,~IEEE,} 
	Aaron~M.~Johnson,~\IEEEmembership{Member,~IEEE,}        \\
        Evan~Chang-Siu,~\IEEEmembership{Member,~IEEE,} 
        Robert~J.~Full,
        and~Daniel~E.~Koditschek,~\IEEEmembership{Fellow,~IEEE}
\thanks{
This work was supported in part by the ARL/GDRS RCTA
and in part by the NSF under the CiBER-IGERT Award DGE-0903711
and the CABiR Award CDI-II 1028237. 
Portions of this work (including some experimental results) previously appeared in~\cite{paper:CLAWAR-Tails}.
}
\thanks{T. Libby is with the Departments of Mechancial Engineering and Integrative Biology, University of California Berkeley,
Berkeley, CA 94720 USA email: tlibby@berkeley.edu.}
\thanks{R.J. Full is with the Department of Integrative Biology, University of California Berkeley,
Berkeley, CA 94720 USA email: rjfull@berkeley.edu.}
\thanks{A. M. Johnson is with the Department of Mechanical Engineering, Carnegie Mellon University, Pittsburgh, PA 15232 USA email:
amj1@andrew.cmu.edu.}
\thanks{E. Chang-Siu is with the Department of Engineering Technology, California State University Maritime Academy,
Vallejo, CA 94590 USA email: echangsiu@csum.edu.}
\thanks{D. E. Koditschek is with the Electrical and Systems Engineering Department, University of Pennsylvania,
Philadelphia, PA 19104 USA email: kod@seas.upenn.edu.}
}

\thispagestyle{empty}
\setcounter{page}{0}
\begin{figure*}[t!]
\centering
\large
This paper has been accepted for publication in IEEE Transactions on Robotics.\\

DOI: \href{https://doi.org/10.1109/TRO.2016.2597316}{10.1109/TRO.2016.2597316}\\

IEEE Explore: \href{http://ieeexplore.ieee.org/document/7562541/}{http://ieeexplore.ieee.org/document/7562541/}\\

~\\

Please cite the paper as:\\

Thomas Libby, Aaron M. Johnson, Evan Chang-Siu, Robert J. Full, and Daniel E. Koditschek, ``Comparative Design, Scaling, and Control of Appendages for Inertial Reorientation,'' in \emph{IEEE Transactions on Robotics}, vol. 32, no. 6, pp. 1380--1398, Dec. 2016.\\

~\\

~\\

\copyrighttext
\vspace{400px}
\end{figure*}
\maketitle
\copyrightnotice
\begin{abstract}

This paper develops a comparative framework for the design of actuated inertial appendages for planar, aerial reorientation.
We define the Inertial Reorientation template, the simplest model of this behavior, and leverage its linear dynamics to
reveal the design constraints linking a task with the body designs capable of completing it. As practicable inertial
appendage designs lead to morphology that is generally more complex, we advance a notion of ``anchoring'' whereby
a judicious choice of physical design in concert with an appropriate control policy yields a system whose closed loop
dynamics are sufficiently captured by the template as to permit all further design to take place in its far simpler
parameter space. This approach is effective and accurate over the diverse design spaces afforded by existing platforms,
enabling performance comparison through the shared task space. We analyze examples from the literature and find
advantages to each body type, but conclude that tails provide the highest potential performance for reasonable designs.
Thus motivated, we build a physical example by retrofitting a tail to a RHex robot and present empirical evidence of its
efficacy.

\end{abstract}

\IEEEpeerreviewmaketitle

\mathchardef\minus="002D

\section{Introduction}
\label{sec:introduction}

\begin{figure}[t]
\begin{center}
\subfigure[]{\label{fig:tailbot}\includegraphics[scale=.393]{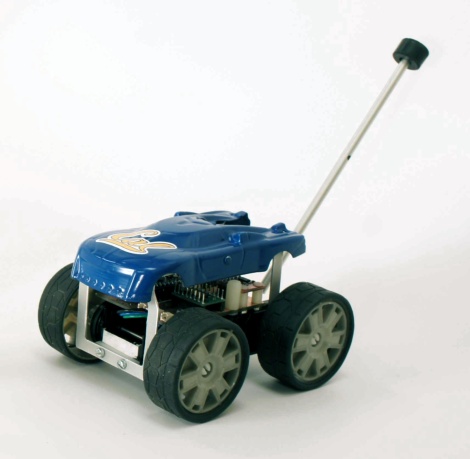}} 
\subfigure[]{\label{fig:xrltail}\includegraphics[scale=.176]{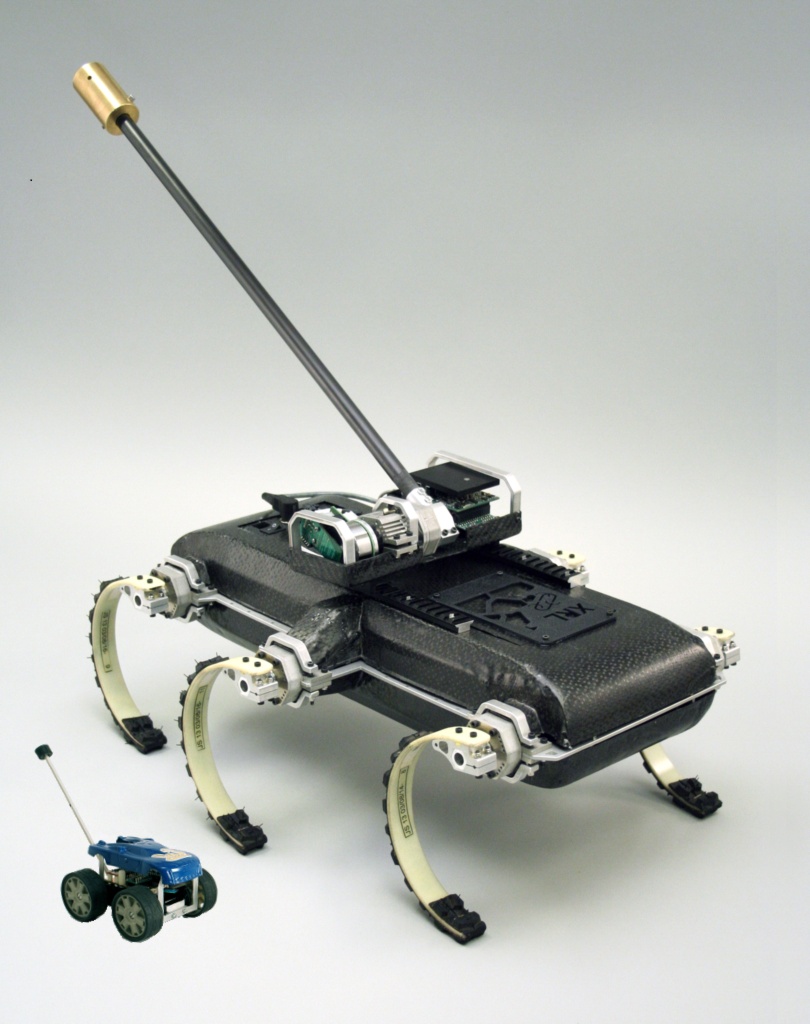}}
\end{center}
\caption{(a) Tailbot~\cite{libby_2012,chang-siu-iros-2011} (b) RHex~\cite{Altendorfer-AR2001,paper:xrhex_canid_spie_2012} with a new tail, and with approximately sized image of Tailbot inserted.
}\label{fig:robots}
\end{figure}

\IEEEPARstart{T}{ails} and tail-like appendages have shown promise to greatly enhance robot agility, 
enabling such feats as aerial reorientation~\cite{chang-siu-iros-2011,paper:CLAWAR-Tails,zhao2015controlling},
hairpin turns~\cite{kohut2013precise,casarez2013using,patel2013rapid}, 
and disturbance rejection~\cite{briggs2012tails,ackerman2013energy,liu2014bio}. 
These behaviors are examples of \emph{Inertial Reorientation} (or \emph{IR}),
whereby internal configuration adjustments generate inertial forces that control the body's orientation. 
The stabilizing function of inertial appendages appears to be important to animals across a wide variety of behaviors and size scales,
suggesting that this mechanism could be broadly useful for robotic systems
such as the small, wheeled Tailbot~\cite{chang-siu-iros-2011}, Fig.~\ref{fig:tailbot},
or the larger, legged RHex~\cite{Altendorfer-AR2001,paper:xrhex_canid_spie_2012},
Fig.~\ref{fig:xrltail}.
While tails may be the most conspicuous example of IR morphology, any internal movement of mass can induce rotation in a body.
Animals also use the inertia of their legs~\cite{pijnappels2010armed, crawford1995biological}, wings~\cite{hedrick2007low}, or spine~\cite{kane1969dynamical}
to accomplish similar behaviors, and engineered systems use radially symmetric wheels inside satellites
or on terrestrial vehicles~\cite{brearley1981motor}.  

This paper presents a formal framework for the selection and comparison of robot 
bodies capable of a planar, aerial, inertial reorientation task.
Design of morphology for a dynamic behavior like IR is a persistently challenging problem in robotics, since 
 task completion
must be enforced over the full design space through the execution of a hybrid and possibly nonlinear dynamical system. We propose a reductionist approach,
collapsing the complexity of the variously possible body plans to a far simpler model whose dynamics we can solve. The task-feasible set of this simple model,
together with its generic controller, is
then pulled back through this ``morphological reduction'' to specify the more complex design.
We use this framework to evaluate the merits of a range of possible morphologies,
and to design a new tail for the RHex robot,
Fig.~\ref{fig:xrltail}, documenting its efficacy for recovery from otherwise  injurious falls as illustrated in Figs.~\ref{fig:rhexfall} and~\ref{fig:rhexcliff}.

\subsection{Prior Work}

The study of inertial reorientation dates to the 19th century ``falling cat problem''~\cite{kane1969dynamical}.
More recent studies show that by swinging their tails, lizards can 
self-right in less than a body length~\cite{jusufi2008active}, reorient through zero net angular 
momentum IR maneuvers~\cite{jusufi_2010}, and control their attitude in leaps~\cite{libby_2012}. 
To the authors' best knowledge, the first robot to utilize an inertial tail is the Uniroo, a one leg hopper that stabilized
its body pitch in part with an actuated tail~\cite{zeglin1991uniroo}. Other early robotic tails were 
passive or slowly actuated and used to maintain contact forces while climbing
vertical surfaces~\cite{autumn2005robotics,kim2005spinybotii}. The idea of using a robot's existing
limbs as tail-like appendages was first explored as a method of ``legless 
locomotion''~\cite{Balasubramanian_2008_6145}. 

The effectiveness of the IR capabilities in lizards inspired the creation of Tailbot, Fig.~\ref{fig:tailbot}, a robot with an active tail 
which enabled disturbance regulation~\cite{libby_2012}, air-righting, and traversing rough terrain~\cite{chang-siu-iros-2011}.  
Since Tailbot, 
there has been an explosion in the number of robotic tails for 
reorientation~\cite{zhao2015controlling,demir2012inertial,kohut2013precise, patel2013rapid, casarez2013using}
and stabilization~\cite{briggs2012tails,ackerman2013energy,liu2014bio,CompositionHoppingTR,paper:heim2015tail} in both aerial and terrestrial domains.
Non-inertial tails have also seen continued interest with tails that
affect the body through substrate interaction \cite{crespi2013salamandra,estrada2014perching,ws:kessens-tail-2015} and aerodynamics~\cite{jusufi2011aerial,ws:patel-tail-2015}.
Recently, other morphologies have also been explored including two degree of freedom
tails that greatly expand the range of possible motions~\cite{2DOFtailbot,liu2014bio,CompositionHoppingTR} and
flailing limbs that reuse existing appendages~\cite{johnson_selfmanip_2013}.
Many of these robots draw their inspiration from a diverse variety of animals, including moths~\cite{demir2012inertial,dyhr2012autostabilizing},
seahorses~\cite{Porter03072015}, 
 kangaroos~\cite{liu2014bio},
cheetahs~\cite{briggs2012tails,patel2013rapid,paper:heim2015tail},
and even dinosaurs~\cite{libby_2012,grossi2014walking}.
The growing interest in robotic IR appendages
demonstrates the potential benefits of inertial forces 
and motivates the need for truly comparative design methodologies.

\subsection{Paper Outline and Contributions}
To instantiate the appendage design problem, in this work we consider an aerial IR self-righting 
task. For whereas while the machines examined in this paper are nominally terrestrial locomotors, their rapid, dynamic behavior
 includes leaps, falls, and other short aerial phases where their limbs cannot provide control authority through ground reaction forces.
We will restrict motion in both the templates and anchors to a 2D plane, in the absence of external forces.
The task is defined as a finite-time, zero angular momentum reorientation: a rotation of the body configuration 
$\theta_b$ from initial condition $\theta_b(0)=\dot{\theta}_b(0)=0$, to rest at some final angle $\theta_{b,f}$ in a desired time $t_f$. 
That~is,
\begin{align}
\theta_b(t_f)-\theta_{b,f} = 0; \qquad \dot{\theta}_b(t_f) = 0. \label{eq:Sr}
\end{align}
Because any internal motion -- whether a rotation of tails, wheels, limbs, or even body bending -- must yield some 
inertial reorientation in flight, we need a method of directly comparing the performance and design merits of a diverse array 
of potential body structures.
The simplicity of the shared underlying behavior 
is suggestive of a \emph{template}~\cite{Full-JEB-1999}, or simplest model,
whose tractable dynamics yield a compact description of the relationship between morphology and task performance.
We present the IR template (Section~\ref{sec:templatekin}) and solve its simple dynamics (Section~\ref{sec:tempdynamics})
relative to the task~\eqref{eq:Sr} (Section~\ref{sec:templatetask}),
revealing the constraints linking that task with the set of body designs capable of completing it.
We then refine that set by reducing it to the instances where the control and gearing are optimal for the 
assigned task (Section~\ref{sec:templatefreedom}).

\begin{figure}[t]
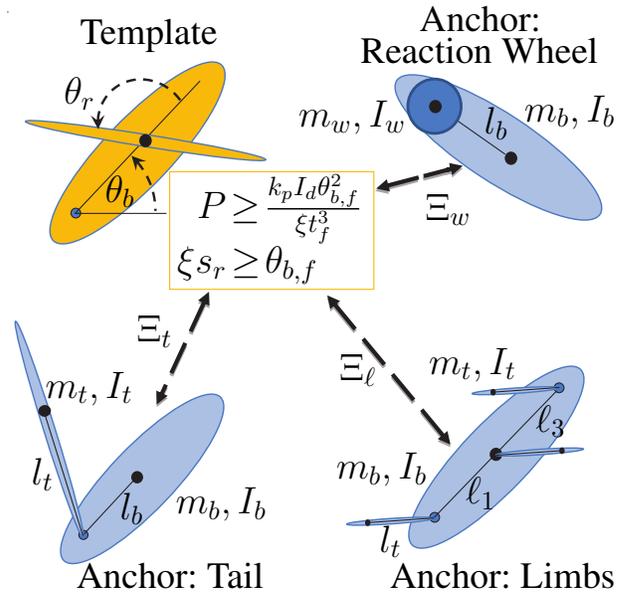

    \centering
    \def\svgwidth{8.5cm}
    \relsize{2}
    \include{TemplateAndAnchors_II}
    \caption{ The Inertial Reorientation Template is a planar, two-link model parametrized in part by Power, $P$, Effectiveness, $\xi$, Appendage Stroke, $s_r$, and Driven Inertia, $I_d$; designs satisfying the constraints are feasible with respect to the task,~\eqref{eq:Sr}.  More complex IR bodies (anchors) 
                   may be designed or compared through the template by mapping their physical parameters to those of the template, using a Morphological Reduction, $\Xi_i$, as summarized in Table~\ref{tab:morpho}.}
    \label{fig:TnA}
\end{figure}

The embodiment of this simple template in a more complex model of real morphology (an \emph{anchor},~\cite{Full-JEB-1999})\footnote{
Here, there is no time-asymptotic specification, and therefore  
no attracting invariant set as achieved empirically, e.g.\ in \cite{Rizzi_Koditschek_1994,Altendorfer-AR2001}, and formally as well, e.g.\ in 
\cite{Nakanishi_Fukuda_Koditschek_2000}.
Instead, we observe that the anchors
manifest a close approximation to the template over large, interesting regions of parameter spaces, Appendix~\ref{app:anchors}.
} 
provides for a shared parametrization of IR efficacy. 
This is a new idea that enables the design and direct comparison of different 
candidate bodies through a generalized template--anchor relationship that we now briefly describe 
intuitively before charting its technical development in this paper and in Appendix~\ref{app:anchors}.
Whereas in this problem the template degrees of freedom typically embed naturally
into those of the morphologies,
the same is not true of their respective design parameters. 
Thus, our agenda of  reusably ``anchoring'' a template design
in a variety of  bodies requires 
a new mapping between their parameter spaces. 
Beyond the specifics of the task, one of
the central contributions of this paper is to articulate and formalize the role of this morphological reduction. 
As we detail in Section~\ref{sec:dynamics}, mapping the design parameters (mass, length, and inertia, etc) of a detailed model down
to the simpler template parameters, carries a pullback of the simple template controller back up to the anchor as well.

We define anchor models for tails (Section~\ref{sec:tailkin}), reaction wheels (Section~\ref{sec:wheels}), 
and synchronized groups of limbs (hereafter termed ``flails'', Section~\ref{sec:kinelimbs}),
and propose morphological reductions from their respective parameter spaces to the parameter space of the template 
(summarized in Fig.~\ref{fig:TnA} and Table~\ref{tab:morpho}).
We use these  morphological reductions to find evidence of 
similar template--anchor relationships in design examples from a dozen different
platforms (Section~\ref{sec:examples}), 
which exhibit close (or in some cases exact) kinematic and dynamic approximations (depicted in Fig.~\ref{fig:Tails_error_fig}).   
The reductions afford a performance comparison of each morphology (Section~\ref{sec:morphology}),  
and a more general comparative scaling analysis (Section~\ref{sec:scaling}).

We assess anecdotally the utility of our design framework in three ways.
First, we use our design specification to analyze the tail added to 
a RHex hexapedal robot~\cite{Altendorfer-AR2001}
(specifically X-RHex Lite, XRL~\cite{paper:xrhex_canid_spie_2012}, Fig.~\ref{fig:xrltail} and Section~\ref{sec:xrltail}).  
Second,
we use the common IR template to compare the RHex tailed-body instantiation with a limbed-body instantiation 
using only RHex's legs (Section~\ref{sec:xrlflail}). 
Finally, we present empirical results (Section~\ref{sec:experiments}) that illustrate the manner in which
IR behaviors can help robots perform high-performance, potentially injurious  aerial
reorientation using inertial limbs.

In the interest of space and clarity, we have omitted the more lengthy derivations required to reach some expressions in this paper;
the full derivations can be found in the accompanying technical report,~\cite{tr:tails-2015}.

\section{Template Behavior}
\label{sec:kinematics}

This section develops the simplest Inertial Reorientation (IR) model and solves its dynamics explicitly in the context of the task specification.
From this, we derive two constraints 
specifying the feasible portion of parameter space over which the robot design
may be optimized  (or -- more practicably  -- ``toleranced,'' as we exemplify in Section V) to best meet performance needs outside the reorientation task. 
To this end, we define the IR template (depicted in Fig.~\ref{fig:TnA}) as a planar system comprised of two rigid 
bodies -- an ``appendage'' and a ``body'' pinned at their shared centers of mass (COMs). 
A motor applies a torque acting on $\theta_r$, the internal angle between the bodies, and can steer $\theta_b$, the orientation of the body,
through the action of a controller.
We will thus choose $(\theta_r,\theta_b)$ as our generalized coordinates. 
The appendage moment of inertia (MOI), $I_a$ and the template body MOI, $I_d$,  
specify the passive mechanics.

The template's behavior during the reorientation task is fully parametrized by a combination of its physical (body) parameters, 
powertrain and control parameters, and its task specification, defined throughout the rest of this section and summarized in Table~\ref{tab:Symbols} as,
\begin{equation}
\label{eq:temp_param}
\mathbf{p} = [\xi, I_d, s_r, P, \omega_m, t_s, \theta_{b,f}, t_f ] \in \mathcal{P}.
\end{equation}
Not all parameter sets $\mathbf{p}$ are self-consistent, as clearly only certain bodies are capable of completing a given task. The remainder of this section will be dedicated to finding 
a parametrization of the constraints defining the feasible subset of parameters, $\mathcal{R} \subset \mathcal{P}$. Any parameter set in $\mathcal{R}$ is ``task-worthy'' in the sense
that its physical parameters enable completion of its task description.
The ``task-worthy'' set will be used to solve two design problems:
\begin{enumerate}[label=P\arabic{*}, ref=P\arabic{*}]
\item \emph{Body Selection}: The task specification is fixed at the outset and the other parameters are chosen to satisfy its completion.
\label{problem:body}
\item  \emph{Performance Evaluation}: The physical parameters are fixed and a given $t_f$ and $\theta_{b,f}$ are queried against a resulting feasible set.
\label{problem:performance}
\end{enumerate}
We next derive the kinematics and dynamics of this IR template model,
and then solve those dynamics in normalized form to reveal the feasible set $\mathcal{R}$. 

\begin{table}[t]
  \begin{center}
    \begin{tabular}{ l l}
      \toprule
      \midrule
      $g_h, g_\theta$ & Time and angle functions \eqref{eq:phys_th}, \eqref{eq:phys_gh}\\
      $\tilde{g}_h, \tilde{g}_\theta, \tilde{g}_c$ & Normalized time and angle functions \eqref{eq:g_h}, \eqref{eq:g_theta}, \eqref{eq:switchcon}\\
      $H_O$ & Angular momentum \eqref{eq:Hodef}\\
      $I_a,I_b,I_t$ & Inertia of the appendage, body, and tail (\ref{sec:templatekin}), (\ref{sec:tailkin})\\
      $I_d,I_{d,t}$ & Driven inertia of the template and tail (\ref{sec:tempdynamics}), \eqref{eq:Idt}--\eqref{eq:idave}\\
      $l_b,l_t$ & Length from the pivot to the body and tail (\ref{sec:tailkin})\\
      $k_p, k_t, k_s$ & Power, time, and speed constants \eqref{eq:kpdef}, (\ref{sec:optgear}) \\
      $\ell_i$ & Limb offsets (\ref{sec:kinelimbs})\\
      $L$ & Characteristic body length (\ref{sec:scaling})\\
      $m_b,m_t$ & Mass of the body and tail (\ref{sec:tailkin})\\
      $m_r$& Reduced mass \eqref{eq:mrdef}\\
      $N$ & Number of limbs (\ref{sec:kinelimbs})\\
      $\mathbf{p} \in \mathcal{P}$ & Template parameters \eqref{eq:temp_param}\\
      $\mathbf{p}_i \in \mathcal{P}_i$ & Anchor $i$ parameters \eqref{eq:p_t}, \eqref{eq:p_w}, \eqref{eq:p_ell}\\
      $P$ & Motor power (\ref{sec:tempdynamics})\\
      $\mathcal{R},\mathcal{R}^*,\mathcal{R}_i$ & Allowable parameter set \eqref{eq:TemplateEvalConstraints}, \eqref{eq:TemplateConstraints}, \eqref{eq:pullback}\\
      $s_r$ & Range of motion (\ref{sec:templatekin}) \\
      $t, t_s, t_h, t_f$ & Time, switching, halting and final time (\ref{sec:templatekin}), (\ref{sec:reduced})\\
      $\tilde{t},\tilde{t}_s,\tilde{t}_c, \tilde{t}_h$ & Normalized, switching, critical, and halting time (\ref{sec:reduced}) \\
      $\gamma$ & Time scaling parameter \eqref{eq:gamma}\\
      $\eta$ & Nonlinearity parameter \eqref{eq:eta}\\
      $\theta_b, \theta_t, \theta_r, \theta_h\!$ & Body, tail, relative, and halting appendage angles (\ref{sec:templatekin})\\
      $\tilde{\theta}$ & Normalized relative angle (\ref{sec:reduced})\\
      $\xi, \xi_t$ & Effectiveness of the template and tail \eqref{eq:xidef}, \eqref{eq:xit}\\
      $\xi_w, \xi_l$ & Effectiveness of the reaction wheel, and limbs \eqref{eq:xiw}, \eqref{eq:xileg}\\
      $\Xi_i$ & Morphological reduction $i$ \eqref{eq:morphred}\\
      $\tau$ & Motor torque (\ref{sec:tempdynamics})\\  
      $\omega_m, \tilde{\omega}_m$ & Motor and normalized no-load speed (\ref{sec:tempdynamics}), (\ref{sec:reduced})\\
      \bottomrule
    \end{tabular}
     \caption{Key symbols used throughout this paper with section or equation number of introduction marked. 
     }
     \vspace{-6ex}
     \label{tab:Symbols}
  \end{center}
\end{table}

\subsection{Template Kinematics}
\label{sec:templatekin}

For a planar, single degree of freedom IR system in free fall, the rotation available in the body's workspace is limited by the capacity for
internal motion.
To derive a functional relationship between the (internal) shape angular velocity and the 
(external) body orientation velocity, we will use the
non-holonomic constraint resulting from conservation of the system's total 
angular momentum. From any point $O$, Euler's laws for a rigid body state that
$\mathbf{\dot{H}_O} = \mathbf{M_O},$
where $\mathbf{H}_O$ is the total angular momentum about $O$, and $\mathbf{M_O}$ is the net moment about $O$.
For short aerial behaviors in robots larger than a few grams, we will assume that the external forces
and torques (particularly aerodynamic torques) are negligible so that $\mathbf{M_O} = 0$, and hence 
total angular momentum about $O$ is conserved.

The template's
angular momentum about the perpendicular axis ($\mathbf{E}_3$) of its~COM, 
\begin{equation}
H_O = (I_a+I_d) \dot{\theta}_b + I_a \dot{\theta}_r, \label{eq:Hodef}
\end{equation}
where $H_O \mathbf{E}_3 := \mathbf{H}_O$ and $\dot{\theta}_b$ and $\dot{\theta}_r$ are derivatives with respect to time $t$.
Normalizing by the total MOI, $I_a+I_d$, and solving for body angular velocity reveals that the template kinematics
are parametrized by a single dimensionless constant $\xi$, the \emph{effectiveness}
of the IR template\footnote{Note
that this quantity differs from that of~\cite{paper:CLAWAR-Tails}, wherein effectiveness $\varepsilon$ was defined as the ratio of link velocities.},
\begin{equation}
\dot{\theta}_b  = \widetilde{H}_O - \xi \dot{\theta}_r, \qquad \xi:=\frac{I_a}{I_a+I_d}, \label{eq:xidef}
\end{equation}
where $\widetilde{H}_O$ is the normalized system angular momentum.  
Hence the angular velocity of the body can be decomposed into
two physically interesting components: a drift term influenced solely by external impulses, 
and the velocity induced by internal shape change that has been called the \emph{local connection vector field} 
\cite{hatton2011geometric} (hereafter \emph{connection field}, 
although note that in this transient setting there is no cyclic shape change).
This equation directly governs performance in two distinct tasks: 
1) orientation regulation after an impulse, where the task is to maintain a stable body 
angle ($\dot{\theta}_b=0$), with a relative velocity $\dot{\theta}_r =  \widetilde{H}_O/\xi$; and 
2) zero angular momentum reorientation ($H_O = 0$), where the task is to change the body
orientation to some angle $\theta_{b,f}$ in $t_f$ seconds,~\eqref{eq:Sr},
given the constraint of the connection field, $\dot{\theta}_b=- \xi \dot{\theta}_r$.

In the latter case, body rotation is directly a function of appendage rotation.
Under the assumption that $\dot \theta_r$ is positive, 
\begin{align}
\dot{\theta}_b = \frac{d \theta_b}{d t} = \frac{\partial \theta_b}{\partial \theta_r} \frac{d \theta_r}{d t} 
&=
 -\xi \dot{\theta}_r, 
\qquad \frac{\partial \theta_b}{\partial \theta_r} = -\xi, \label{eq:templatechain}
\end{align}
expressing the 1-dimensional connection field that reveals the constant 
differential relationship between internal and external rotation.\footnote{In the anchor models this relationship may be nonlinear or non-monotonic.}
For this template, the connection field is constant and equal to $-\xi$. The body stroke is directly 
proportional to appendage stroke, and hence a limit $s_r$ on the range of motion of the appendage will limit the achievable body rotation.

\subsection{Template Dynamics}
\label{sec:tempdynamics}
A real terrestrial robot is constrained by the duration
of its aerial phase (fall, leap, or other dynamic behavior) and this imposes
a new set of requirements on the parameters that specify the actuation.
This section characterizes the behavior of a conventionally power-limited actuation scheme,
and defines a controller for that actuator.

\subsubsection{Newtonian and Actuator Dynamics}
As the template consists of two rigid bodies pinned through their concentric COMs, derivation of the equations of motion is trivial -- 
the angular acceleration of body and tail are opposite in sign and equal to the motor torque normalized by each body's MOI.
Since the tail angle is kinematically related to that of the body by \eqref{eq:templatechain}, we will simply consider the body~dynamics,
\begin{equation}
\label{eq:kappatau}
\ddot{\theta}_b = \frac{\tau}{I_d},
\end{equation}
where $\tau$ is the motor torque. 
The ratio of joint torque to body angular acceleration is equal to the body's MOI in the template,
$I_d$, but 
is more complex in the anchors (coupling appendage masses, etc.); to avoid confusion with 
the inertia of the physical body segment in the anchor models, 
we will call this ratio the ``driven'' inertia.

To capture the essential limitation of any powertrain in a time-sensitive task -- the rate at which it can change the mechanical 
energy of the driven system -- we augment the template's dynamics with a simple, piecewise-linear actuator model in which torque
falls linearly with increasing speed (we extend this to allow for current limits in Appendix~\ref{app:cl}). 
This model is not only a good approximation of a DC motor~\cite{Gregorio_Ahmadi_Buehler_1997}, but 
is general enough to capture to first order the effort-flow relationships of many other
speed-dependent actuators including biological muscles~\cite{hannaford1990actuator}.
The maximum available actuator torque 
depends on activation (terminal voltage, $V=\pm V_m$, for some maximum
voltage $V_m$) and~speed,
\begin{equation}
\tau(V, \dot{\theta}_r) = \left\{
  \begin{array}{lr}
   \sgn{(V)}~ \tau_m \left(1-\frac{|\dot{\theta}_r|}{\omega_m}\right)  & : V \dot{\theta}_r < 0 \\
   \sgn{(V)}~ \tau_m  & : V \dot{\theta}_r \ge 0
  \end{array}
\right. \label{eq:motor}
\end{equation}
where $\tau_m$ is the stall torque and $\omega_m$ is the no-load speed of the motor after the gearbox 
(and hence the no-load speed of the appendage relative to the body). 

Since we seek to specify the entire powertrain, we find it convenient to 
decouple the roles of the actuator and the transmission  
by parametrization with respect to peak mechanical power, 
$P=\tau_m\omega_m/4$,  (whose product form cancels the appearance of the gear ratio) 
and drivetrain no-load speed, $\omega_m$ (whose linear dependence upon the gear ratio makes it a useful surrogate for the transmission).
The required gear ratio of a physical gearbox or other transmission is then the ratio of $\omega_m$ to the motor's actual 
no-load speed.

\subsubsection{Controller Design}

Notwithstanding the voluminous literature on time optimal control in mechatronics and robotics settings (e.g., along specified paths \cite{Bobrow_Dubowsky_Gibson_1985},
and exposing actuator dynamics \cite{Tarkiainen_Shiller_1993}) we have not been able to find a formal 
treatment of the robust minimum time problem for our simple hybrid motor
model~\eqref{eq:motor}. Therefore we will take the na\"{i}ve approach and embrace a single switch open loop bang-bang 
controller as offering the simplest and most paradigmatic expression of ``fast repositioning'' 
for a (back-EMF perturbed) double integrator \cite{Rao_Bernstein_2001}. 
We relax the bang-bang controller assumption in Appendix~\ref{app:altcont} and in particular show that a proportional-derivative (PD)
feedback controller closely and robustly approximates (and given high enough gains, converges to) the open loop control policy. 
We further verify this in the empirical results, Section~\ref{sec:experiments}, which use a
PD controller to approximate the bang-bang~controller.
 
The bang-bang control strategy makes a single switch between the acceleration and braking dynamics 
at a time $t_s$, such that the body comes to a halt at the desired final orientation $\theta_{b,f}$.
\footnote{This may be replaced by an event-based guard condition $G({\theta}_b, \dot{\theta}_b) = 0$,
as derived in~\trsecref{trapp:event}.}
During the single-switch reorientation from $\theta_b = 0$, the body will accelerate from rest and brake to the final angle $\theta_b = \theta_{b,f}$ with no overshoot, with
$\dot{\theta}_b \geq 0$ and $\dot{\theta}_r \leq 0$ for the entire maneuver.  
Using~\eqref{eq:templatechain}, the torque can be rewritten 
to eliminate the dependence on $\theta_r$.
The hybrid dynamics are described by an
acceleration phase and a braking phase,
\begin{align}
\ddot{\theta}_b = \left\{ 
\begin{array}{ll}
\displaystyle
\frac{4{P}}{\omega_m I_d}\left(1-\frac{\dot{\theta}_b}{\xi \omega_m}\right), \qquad &\mathrm{for} ~ 0\leq t<t_s, \\
\displaystyle
-\frac{4{P}}{\omega_m I_d}, &\mathrm{for} ~ t\geq t_s. 
\end{array}\right.
\label{eq:ddthetab}
\end{align}

\subsubsection{Behavior in reorientation task}
\label{sec:templatebehavior}

Based on this template kinematics, dynamics, and controller structure, we now examine the resulting behavior of the
system in this reorientation task.
First, note that due to local integrability of the non-holonomic constraint,~\eqref{eq:templatechain}, the system has only a single degree of freedom after the initial conditions 
are chosen.  We therefore choose to define the initial conditions as $\theta_r=\theta_b:=0$, and express the dynamics
only in terms of $\theta_b$. The system starts at rest, so that $\dot{\theta}_b = 0$. 
We can write the system behavior in closed form by integrating the linear, switched dynamics in \eqref{eq:ddthetab} from this initial condition until the body again comes to a halt at a time $t_h$.
See \trsecref{tr:optimtemp} for details on this integration. 
The halting time can be written as an explicit function of the template parameters,~\eqref{eq:temp_param},
\begin{equation}
\label{eq:phys_th}
t_h = g_h(\mathbf{p}) := t_s + \frac{I_d \xi \omega_m^2}{4 P} \left(\!1 - \exp\left(\! -\frac{4 P}{ I_d \xi \omega_m^2} t_s\!\right)\!\right),
\end{equation}
along with the final angle, $\theta_b = \theta_h$,
\begin{equation}
\label{eq:phys_gh}
\theta_h = g_\theta(\mathbf{p}) := \xi \omega_m t_s - \frac{I_d\xi^2 \omega_m^3}{8 P} \left(\!1 - \exp\left(\! -\frac{8 P}{ I_d \xi \omega_m^2} t_s\!\right)\!\right).
\end{equation}

\subsection{Dynamical Task Encoding}
\label{sec:templatetask}

The physical relationships derived in the previous two sections enable a straightforward representation of the task-feasible parameter subset $\mathcal{R}$
containing all self-consistent parameter sets. 
This restricted set can be written as a system of constraints to facilitate the two design problems identified at the beginning of this section: \ref{problem:body} \emph{Body Selection},
in which the task specification ($t_f$ and $\theta_{b,f}$) is fixed at the outset and $\mathcal{R}$ 
prescribes the corresponding feasible body designs, and \ref{problem:performance} \emph{Performance Evaluation}, where the
achievable task set is identified, given a fixed body design (values of $\xi$, $I_d$, $P$, $\omega_m$, $s_r$, and $t_s$).

The first constraint arises from the kinematic relation,~\eqref{eq:templatechain}, and ensures that the rotation by the task, $\theta_{b,f}$ falls within any physical constraints on rotation.
If the design has a finite range of motion $s_r$ (so that $\theta_r \in [0, s_r]$), then any design meeting the task specification~\eqref{eq:Sr} must satisfy,
\begin{equation}
\label{eq:proptoxi}
\xi s_r \geq \theta_{b,f};
\end{equation}
obviously bodies with unlimited range of motion satisfy this constraint trivially.  
The second constraint ensures that the halting time,~\eqref{eq:phys_th}, falls within the task completion time, $t_f$. 
The third constraint ensures that the body, under the bang-bang controller
(parametrized by $t_s$), \eqref{eq:phys_gh}, stops at the correct angle. Taken together, these constraints define $\mathcal{R}$,
\begin{align}
\label{eq:TemplateEvalConstraints}
\mathcal{R} :=&  \Big\{ \mathbf{p} \in \mathcal{P} \, \Big| \,
\xi s_r \geq \theta_{b,f}, \,
t_f \geq g_h(\mathbf{p}), \,
\theta_{b,f} = g_\theta(\mathbf{p})
\Big\}.
\end{align}
For the \emph{Body Selection} problem, \ref{problem:body}, any design, $\mathbf{p}\in \mathcal{R}$, satisfying these constraints is 
``task-worthy'' in that its physical and controller parameters satisfy its task specification. The \emph{Performance Evaluation} 
problem, \ref{problem:performance}, is also easily specified using this representation: fixing all parameters save $t_f$ and $\theta_{b,f}$ 
specifies a two-dimensional subspace of achievable tasks (see Fig.~\ref{fig:taskspace}
for a graphical example).\footnote{The largest task set will be found  by 
allowing the switching time to vary with the task (i.e., using the third constraint in~\eqref{eq:TemplateEvalConstraints} to select $t_s$ for each $\theta_{b,f}$.)}

Unfortunately, $\mathcal{R}$ still leaves many degrees of freedom for task-worthy designs for the \emph{Body Selection} design problem. 
In the remainder of this section, we show that the gearing and control parameters ($\omega_m$ and $t_s$, respectively) can be eliminated through optimization, thereby enabling a more compact
and considerably more prescriptive set. 

\subsubsection{Spatiotemporally-normalized template behavior}
\label{sec:reduced}

The isolation of the effect of gearing and control on $\mathcal{R}$ is complicated by their nonlinear interaction with the other dimensioned parameters in $\mathbf{p}$. To remove
the effect of scale and expose these relationships, we will nondimensionalize the equations~\eqref{eq:phys_th} and~\eqref{eq:phys_gh},
seeking a spatiotemporal rescaling\footnote{This rescaling can also be seen as a nondimensionalization of the template dynamics resulting in a normalized hybrid system that simplifies the integration of the dynamics; see~\trsecref{tr:optimtemp}.}
parametrized by $\gamma$, such~that,
\begin{equation}
\label{eq:spatiotemporal}
\tilde{t}_s = \gamma t_s; \quad \tilde{t}_f = \gamma t_f; \quad \tilde{t}_h = \gamma t_h; \quad \tilde{\theta}_h = \frac{\theta_h}{\theta_{b,f}}.
\end{equation}
where the$~\tilde{\cdot}$
indicates dimensionless values. We find that choosing,
\begin{equation}
\gamma := \left(\frac{4{P}\xi}{I_d \theta_{b,f}^2}\right)^{\frac{1}{3}}, \label{eq:gamma}
\end{equation}
enables a particularly convenient reduction of $g_h$ and $g_\theta$, \eqref{eq:phys_th}--\eqref{eq:phys_gh}, 
written as a function of only two normalized parameters,
\begin{align}
 \tilde{t}_h &=\tilde{g}_h(\tilde{\omega}_m, \tilde{t}_s) 
 :=  \tilde{t}_s + \tilde{\omega}_m^2 \left(1-\exp\left(\frac{-\tilde{t}_s}{\tilde{\omega}_m^2}\right)\!\!\right) 
 \label{eq:g_h}\\
\tilde{\theta}_h &=\tilde{g}_\theta(\tilde{\omega}_m, \tilde{t}_s) :=  \tilde{\omega}_m \tilde{t}_s -\frac{\tilde{\omega}_m^3}{2}\left(1-\exp\left(\frac{-2\tilde{t}_s}{\tilde{\omega}_m^2}\right)\!\!\right), \label{eq:g_theta}
\end{align}
where $\tilde{\omega}_m$ is a dimensionless actuator parameter that stands as a proxy for gearing,
\begin{equation}
\label{eq:tildeomega}
\tilde{\omega}_m :=\frac{\xi \omega_m}{\gamma \theta_{b,f}}.
\end{equation}

In the rescaled coordinates, the reorientation task requires that the system halt at $\tilde{\theta}_h = 1$, constraining the normalized parameters to one degree of freedom.
This freedom can be parametrized by $\tilde{\omega}_m$ through the implicit function specifying
the ``critical'' switching time $\tilde{t}_c$, satisfying $\tilde{g}_{\theta}(\tilde{t}_s, \tilde{w}_m) = 1$ for a given choice of no-load speed,
\begin{equation}
\label{eq:switchcon}
\tilde{t}_c = \tilde{g}_c(\tilde{\omega}_m) := \inf\{\tilde{t}_s > 0 ~|~ \tilde{g}_\theta(\tilde{t}_s,\tilde{\omega}_m) = 1 \}. 
\end{equation}
When the other system parameters are chosen, the designer can  choose the controller that completes the task by setting
\begin{equation}
t_s = \gamma \tilde{g}_c\left(\frac{\xi \omega_m}{\gamma \theta_{b,f}}\right),
\end{equation}
automatically satisfying (and therefore obviating the need for) the third constraint in~\eqref{eq:TemplateEvalConstraints}.
With this choice, the scaled halting time depends only on the scaled no-load speed,
\begin{equation}
\tilde{t}_h = \tilde{g}_h(\tilde{\omega}_m, \tilde{g}_c(\tilde{\omega}_m)).
\end{equation}

The second constraint in~\eqref{eq:TemplateEvalConstraints} can now be written in a more useful form.
The temporal demands of the task require that full template parameters,~\eqref{eq:temp_param},  be chosen so that the 
spatiotemporal rescaling meets the task specification.
In particular, the value of $\gamma$,~\eqref{eq:gamma} (chosen through the selection of physical parameters) must
ensure that the physical halting time meets the constraint,
\begin{align}
t_f \geq t_h = \frac{1}{\gamma} \tilde{t}_h = \frac{1}{\gamma}  \tilde{g}_h(\tilde{\omega}_m, \tilde{g}_c(\tilde{\omega}_m)).
\label{eq:timeconst}
\end{align}

Substituting the definition of $\gamma$ and rearranging terms yields a more compact version of the time constraint in~\eqref{eq:TemplateEvalConstraints}, predicated on 
critical switching time,
\begin{align}
\frac{\xi P}{I_d} &\geq k_p \frac{\theta_{b,f}^2}{t_f^{3}}, \label{eq:poweropt} 
\end{align}
where $k_p$ is a function of dimensionless gear ratio defined as,
\begin{equation}
\label{eq:kpdef}
k_p:= \frac{1}{4}\tilde{g}_h^3\left(\frac{\xi \omega_m}{\gamma \theta_{b,f}}, \tilde{g}_c \left(\frac{\xi \omega_m}{\gamma \theta_{b,f}}\right)\right).
\end{equation}
For a fixed task specification with a given inertia, power and effectiveness trade off directly.
The value of $k_p$ increases the requirements on $P$ and $\xi$, and thus $k_p$ may be considered a performance ``cost''
imposed by suboptimal gearing. We will consider this cost when selecting an actuator design for RHex in Section~\ref{sec:motorsizing}.

\subsubsection{Optimal Control and Gearing for the Template}
\label{sec:optgear}

\begin{figure}
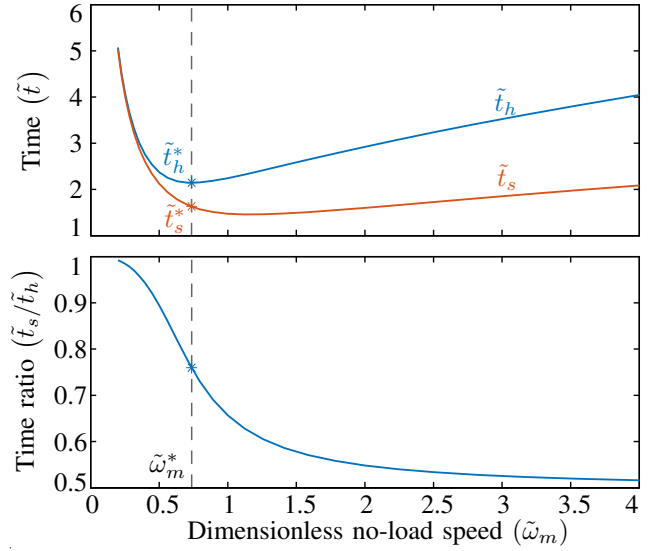

    \centering
    \def\svgwidth{8.5cm}
    \include{GearOpt}
    \caption{Dimensionless system dynamics.  Final time is globally minimized by $\tilde{\omega}_m \approx 0.74$ (top).
		Bang-bang control depends on $\tilde{\omega}_m$; at minimum final time, voltage switches at $\approx 76\%$ of final time.}
    \label{fig:nd_opt}
\end{figure}

The gearing that maximizes performance in
the critically-switched task minimizes $k_p$, or equivalently, the dimensionless completion time $t_h$,
\begin{equation}
\begin{aligned}
& \underset{\tilde{\omega}_m}{\text{minimize}}
& & \tilde{t}_h = \tilde{g}_h(\tilde{\omega}_m, \tilde{g}_c(\tilde{\omega}_m)) .
\end{aligned}\label{eq:mintfun}
\end{equation}
This problem has a (numerically determined) unique global minimum at,
\begin{equation}\label{eq:DimensionlessOptimum}
\tilde{\omega}^*_m \approx 0.74,
\end{equation}
corresponding to a minimal final dimensionless time, $\tilde{t}^*_h:=\tilde{g}_h(\tilde{g}_c(\tilde{\omega}_m^*),\tilde{\omega}_m^*) \approx 2.14$ (Fig.~\ref{fig:nd_opt}, top).
With this optimal $\tilde{\omega}^*_m$ we can find the minimal $k_p^*:=\tilde{g}_h^3(\tilde{\omega}^*_m, \tilde{g}_c(\tilde{\omega}_m^*))/4  \approx 2.46$,
corresponding to the minimal power requirement for \eqref{eq:poweropt}.
Similarly, the critical switching time at this optimum,~\eqref{eq:switchcon}, is a constant
$k_t^* := \tilde{g}_c(\omega_m^*) \approx 1.62$. Finally, the optimal dimensioned no-load speed, $\omega_m$, can
be found from  equations~\eqref{eq:timeconst}
and~\eqref{eq:tildeomega}, ${\omega}_m = k_s \theta_{b,f}/\xi t_f$, for 
$k_s :=  \tilde{\omega}_m\ \tilde{g}_h(\tilde{\omega}_m,\tilde{g}_c(\tilde{\omega}_m))$ (where with
these optimal values, $k_s^* \approx1.58$).

\begin{figure}
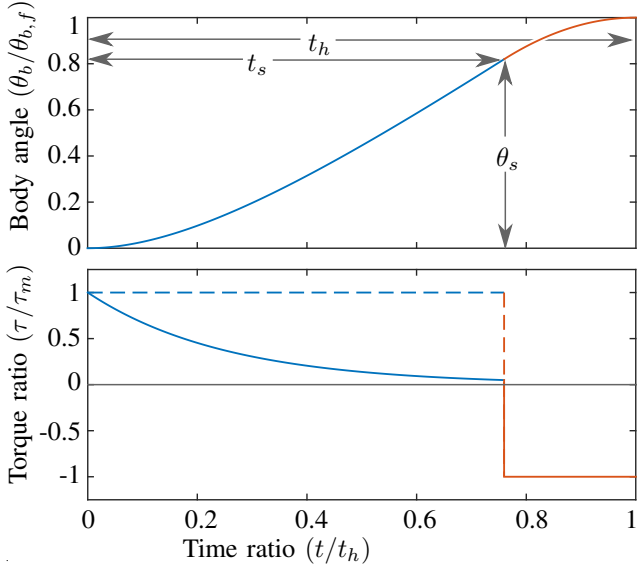

    \centering
    \def\svgwidth{8.5cm}
    \include{control}
    \caption{System kinematics (top). Bang-bang control for optimal gearing (bottom) selects maximal forward input (dashed line) for 76\% of final time, 
     then switches to full reverse; actual torque (solid line) is limited by back-EMF during acceleration (blue) and current during braking (red).     
     }
    \label{fig:control}
\end{figure}

This optimal bang-bang control can be expressed via the ratio $\tilde{t}_s/\tilde{t}_h$
(Fig.~\ref{fig:nd_opt}, bottom); the optimized maneuver 
consists of full positive voltage for $76\%$ of the total time, followed by 
full negative voltage until the body comes to a halt (Fig.~\ref{fig:control}).

The designer seeking the optimally-geared body for a critically-switched reorientation task can then
consider a refinement to $\mathcal{R}$, \eqref{eq:TemplateEvalConstraints}, that
explicitly slaves two of the parameters ($\omega_m$ and $t_s$) to the others,
\begin{align}
\label{eq:TemplateConstraints}
\mathcal{R}^* :=  \Big\{ \mathbf{p} \in \mathcal{P} \, \Big| \,
&\xi s_r \geq \theta_{b,f},\quad
\frac{\xi P }{I_d} \geq \frac{{k}_p^* \theta_{b,f}^2}{t_f^3}, \\
&\omega_m = k_s^* \frac{\theta_{b,f}}{\xi t_f }, \quad
t_s = k_t^* \left(\frac{4{P}\xi}{I_d \theta_{b,f}^2}\right)^{\frac{1}{3}}
\Big\}. \notag
\end{align}

\subsection{Summary of template design freedom}
\label{sec:templatefreedom}

The solution of the template's kinematics and dynamics enabled two representations of the task-feasible subset of design parameters, each serving a particular role in
the two design problems specified at the beginning of this section. Starting with a fixed task specification 
($t_s$ and $\theta_{b,f}$), the \emph{Body Selection} problem, \ref{problem:body}, can be summarized as a choice of
the body parameters ($\xi$, $I_d$, $s_r$, and $P$) subject to the set constraint $\mathcal{R}^*$,~\eqref{eq:TemplateConstraints}, 
with the control and gearing ($\omega_m$ and $t_s$) selected optimally based on this design. 
Alternatively, given an existing (or putative) design, the set $\mathcal{R}$,~\eqref{eq:TemplateEvalConstraints},
can be used in a \emph{Performance Evaluation} problem, \ref{problem:performance},
specifying the achievable tasks. In this latter case, the ``cost'' of suboptimality can be computed 
using $k_p$,~\eqref{eq:kpdef}, or by finding an empirical $k_p$ by substituting the template parameters
into~\eqref{eq:poweropt}.\footnote{A submaximal limit on torque, or suboptimal controllers like the PD scheme discussed earlier, also manifest as an
increase in $k_p$.}

\section{Anchoring via Morphological Reduction}
\label{sec:dynamics}

\begin{table*}[tb]
  \centering
  	\caption{Morphological Reductions for Three Candidate Anchoring Bodies}
	  \label{tab:morpho}
    \begin{tabular}{ l c c c c c c c c c c c}
      \toprule
      Attribute & Tail & Reaction wheel  & Limbs \\
      \midrule
      Inertial Effectiveness, ${\Xi_{i,\xi}}$ & $\frac{I_t + m_r l_t^2}{I_t + I_b + m_r(l_t^2 + l_b^2)}$ &$ \frac{I_w}{I_w+I_b + m_r l_b^2} $ &$ \frac{N (I_t + m_k l_t^2) }{I_b  + m_t \sum\limits_{i=1}^N \ell_i^2+N (I_t + m_k l_t^2)}  $\\
      Driven Inertia, ${\Xi_{i,I_d}}$  & $ (I_b+ m_r l_b^2)(1-\frac{2\eta}{\pi})  $&$  I_b + m_r l_b^2 $&$  I_b  + m_t \sum\limits_{i=1}^N \ell_i^2  $\\
      Anchoring accuracy\textsuperscript{\ref{fn:tailap}} &  Approximate & Exact & Exact \\
	    \bottomrule\\	      
    \end{tabular}
    \vspace{-6ex}
\end{table*}

The concentrically-pinned appendage of the template is not likely to exactly model practical physical designs, raising  
the question of how the template parametrization relates to real bodies available to a robot designer.   
We now explore how the task-feasible restriction on template parameters, $\mathcal{R}$ in \eqref{eq:TemplateEvalConstraints} 
(or with optimized gearing and switching time, $\mathcal{R}^*$ in \eqref{eq:TemplateConstraints}) is reflected in 
the physical parameters (length, mass, and inertia) of bodies a designer might select for inertial reorientation.
A particular template instantiation, $\mathbf{p} \in \mathcal{P}$  could be embodied in myriad ways.
This paper considers three categories of physical IR morphologies that have appeared in the literature: tails, radially-symmetric reaction wheels,
and coordinated flailing limbs, with respective design spaces, $\mathcal{P}_t$, $\mathcal{P}_w$, and $\mathcal{P}_\ell$. 
While the physical parameters and dynamics for these systems differ considerably, they all share the same configuration space and (scalar)
control input space.\footnote{The limbed body is, of course, intrinsically possessed of higher DOF. Here we consider only the case where a coordinating controller has 
rendered its input and state spaces identical to the template. See Appendix~\ref{app:anchors} for a full treatment of this anchoring.}
Therefore the state and input spaces can be mapped from template to anchor trivially, and we focus our attention on the problem of the parameter spaces.
In this section, we show that these bodies 
can be put into formal correspondence with the template task representation by the introduction of a
mapping from these spaces to that of the template,
\begin{equation}
{\Xi_i}: \mathcal{P}_i \rightarrow \mathcal{P}, \label{eq:morphred}
\end{equation}
for $i \in \{t,w,\ell\}$, hereafter termed a \emph{morphological reduction}. 

The morphological reduction affords designers of these bodies the same insight achieved for templates.
The ``pullback'' of the feasible set of body and task parameters through these maps yields an anchoring design in the sense of 
guaranteeing task achievement over the entire inverse image,
\begin{equation}
\label{eq:pullback}
\mathbf{p_{i}} \in  \mathcal{R}_i := {\Xi}_i^{-1}({\mathcal{R}})  \subset \mathcal{P}_i,
\end{equation}
(or similarly, $\mathcal{R}_i^* := {\Xi}_i^{-1}({\mathcal{R}^*})$),
The \emph{Body Selection} and \emph{Performance Evaluation} problems of the previous section can be expressed in the anchor's task-feasible space $\mathcal{R}_i$ by fixing
either the task parameters or body parameters, respectively. We will employ both methods to explore reorientation morphology on RHex in Section~\ref{sec:motorsizing}.

The kinematics and dynamics of anchors may deviate from that of the template, introducing nonlinearities and configuration dependence into the relationships
corresponding to those derived in Section~\ref{sec:kinematics}. For these systems, the morphological reduction is an approximation, with error that varies with task
specification and morphology.\footnote{As shown in this section, 
the tail anchoring is exact when $l_b=0$, the wheel anchoring is always exact, and the limb anchoring is exact 
only for the symmetry conditions described in Section~\ref{sec:kinelimbs}. \label{fn:tailap}}

For the physical bodies discussed in this manuscript,
the parameters defining the powertrain ($P$, $\omega_m$, $s_r$), control ($t_s$), and task ($t_f$, $\theta_{b,f}$) have direct correspondences in both the template and anchor design spaces,
and thus those components of $\Xi$ are simply the identity map and we use the same
notation to describe these quantities in both template and anchor.
However, equivalent parameters for effectiveness and inertia are not obvious a priori and 
therefore are the focus of the following sections (as summarized in Table~\ref{tab:morpho} and Fig.~\ref{fig:TnA}).
As shorthand for these non-trivial components of $\Xi$ we use
${\Xi_{i,\xi}}$ and ${\Xi_{i,I_d}}$ to denote the canonical projection of $\Xi_i$ onto $\xi$ and $I_d$, respectively.

\subsection{Tailed Morphological Reduction}
\label{sec:tailkin}

Within this manuscript, we refer to any single mass-offset appendage specialized for 
inertial reorientation as a ``tail''  (in contrast to flywheels and limbs,
described below), though this configuration could also represent a two-segment body with an actuated 
spine~\cite{paper:mather-iros-2009,paper:xrhex_canid_spie_2012}.
As in the template, the tailed system  consists of two rigid bodies and one internal degree of freedom, but in this case the mass centers of the bodies 
are offset from the joint by some distance ($l_b$ and $l_t$, for body and tail, respectively), 
and the derivation of the connection field is considerably more involved. 
The full parameter set for a tailed body motion is,
\begin{align}
\mathbf{p_t}:=[m_b, I_b, l_b, m_t, I_t, l_t, s_r, P, \omega_m, t_s, \theta_{b,f}, t_f], \label{eq:p_t}
\end{align}
that is, mass, inertia, and COM distance from pivot for each of body and tail (Fig.~\ref{fig:TnA}), as well as the appendage stroke, actuator power,
no-load speed, controller switching time, and task specification.

\subsubsection{Tailed Body Kinematics}
\begin{figure*}[t]
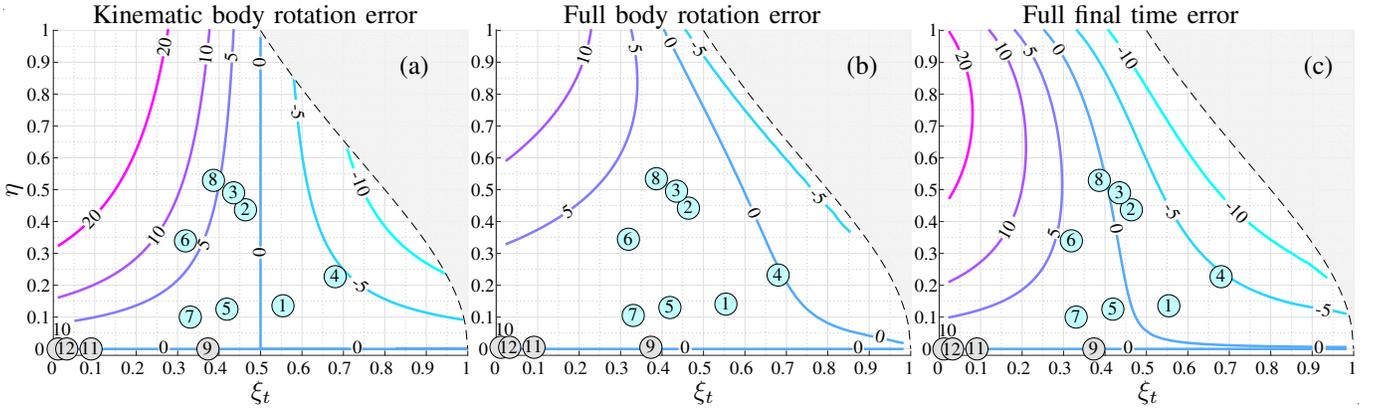

    \centering\def\svgwidth{18cm}
    \scriptsize
    \include{TailsErrorFig}
    \caption{Percent errors of approximation for tailed systems undergoing a half tail rotation centered around $\theta_r=180^\circ$.  
     Numbered points in gray and blue correspond to examples listed in Tables~\ref{tab:properties_limb} \&~\ref{tab:properties}, respectively. Percent error in (a) body rotation due to effectiveness approximation, $\eta=0$ (level sets of~\eqref{eq:errorkin});
    (b) dimensionless final body rotation due to template optimization (level sets of~\eqref{eq:err_body}); and (c) final time due to template optimization
    (level sets of~\eqref{eq:err_time}). Note that for the full body and time error all examples lie within $5\%$ error.}
    \label{fig:Tails_error_fig}
\end{figure*}

The magnitude of the angular momentum about the system COM is nonlinearly configuration-dependent (see~\trsecref{tr:tails} for full derivation), 
\begin{align}
H_{O,t}  &= (I_b + I_t + m_r (l_b^2 + l_t^2 - 2 l_b l_t  \cos{\theta_r})) \dot{\theta}_b \label{eq:magHOt} \\
	& \qquad + (I_t + m_r (l_t^2 - l_b l_t  \cos{\theta_r})) \dot{\theta}_r ,  \notag
\end{align}
where,
\begin{align}
m_r:=\frac{m_b m_t}{(m_b+m_t)} \label{eq:mrdef}
\end{align}
is known as the reduced mass.
As in~\eqref{eq:xidef}, normalize the angular momentum by the total MOI\footnote{The total MOI for a general tail is configuration-dependent;  we take the MOI 
at $\theta_r = \pm 90^\circ$ to achieve the compact form presented here.} about the COM, 
$I_b + I_t + m_r(l_t^2 + l_b^2)$,
and define two dimensionless parameters -- an equivalent~effectiveness, 
\begin{equation}
\xi_t := \frac{I_t + m_r l_t^2}{I_t + I_b + m_r(l_t^2 + l_b^2)}, \label{eq:xit}
\end{equation}
and a nonlinearity parameter,
\begin{equation}
\eta := \frac{m_r l_b l_t}{I_t + m_r l_t^2}. \label{eq:eta}
\end{equation}
The normalized angular momentum is thus,
\begin{align}
\widetilde{H}_{O,t}
&= (1-2\xi_t \eta \cos{\theta_r} ) \dot{\theta}_b + \xi_t (1-\eta \cos{\theta_r} )  \dot{\theta}_r. \label{eq:H_O_hat}
\end{align}
The second dimensionless constant, $\eta$, captures the extent to which the system deviates from the linear behavior
of the template.
Only a subset of the dimensionless parameter space is physically realizable because 
of coupling between the dimensionless constants and the requirement of non-negativity
of the dimensioned parameters (see~\trsecref{app:nondimdom}).
The unreachable region is shaded gray in Fig.~\ref{fig:Tails_error_fig}.

As in~\eqref{eq:templatechain}, setting $\widetilde{H}_{O,t}=0$ and applying the chain rule yields the connection field for the
tail anchor,
\begin{equation}
\frac{\partial \theta_b}{\partial \theta_r}(\theta_r) = -\xi_t \frac{ 1-\eta \cos{\theta_r} }{1-2\xi_t \eta \cos{\theta_r}}.
\label{eq:Atdef}
\end{equation}
Note that ${\partial \theta_b}/{\partial \theta_r} = -\xi_t = const$ when $\eta = 0$ or when $\xi_t = 0.5$, and note that the denominator is 
nonzero when $2\xi_t\eta < 1$, which is always true for physically-realizable parameters (again, see~\trsecref{app:nondimdom}). When $\eta > 1$, the sign of the connection may change
over the tail's range of motion so that transiently both tail and body rotate in the same direction.  

Note that the kinematics are completely described by the connection field, and so two systems with the same $\xi_t$ and
$\eta$ have equal rotations of the body for any given tail rotation. Thus tradeoffs in the physical parameters
$(m_b,I_b,l_b,m_t,I_t,l_t)$ that leave the dimensionless parameters $(\xi_t,\eta)$ unchanged have no effect
on the kinematics of the system.  
In terms of the physical parameters of a robot and tail, this 1-dimensional connection field is,
\begin{equation}
\label{eq:A_nonlin}
\frac{\partial \theta_b}{\partial \theta_r}(\theta_r) =-\frac{I_t + m_r(l_t^2 - l_b l_t \cos{\theta_r})}{I_b+I_t + m_r(l_t^2+l_b^2 - 2 l_b l_t \cos{\theta_r})}. 
\end{equation}
This quantity is at most unity (when the tail is infinitely long or heavy), and
varies over both the configuration space of the robot and its design space.

For tails pivoting directly at the 
body COM, $l_b=0$, the nonlinear terms vanish as $\eta=0$, and the tail anchors to the template without error via
equivalent effectiveness  $\xi_t$.  In general, the connection is not constant and the anchoring is approximate;
this can be accomplished in a number of ways.  The simplest approach (used for the rest of this paper) is to 
assume negligible effect of nonlinearity, i.e.\ $\eta\approx0$, and simply choose
${\Xi_{t,\xi}}(\mathbf{p_t}) := \xi_t$
as in the body-centered case.  
This choice of (approximate) morphological reduction is not unique, and may not be the most accurate in all situations, 
but it works well for all tailed robots described in Table~\ref{tab:properties}.
One alternative is to assume a small range of motion and evaluate the connection field at an intermediate value, such as 
${\Xi_{t,\xi}}(\mathbf{p_t})=  {\partial \theta_b}/{\partial \theta_r}(180^\circ) = \xi_t(1+\eta)/(1+2\xi_t\eta)$.
The most accurate approximation for large tail swings is the average value over the full tail stroke
(which can be found by integration of the connection field as shown in~\trsecref{tr:kinematics}).
This can be found in closed form but the equation's complexity makes it cumbersome 
as a design tool, though useful for calculating or reducing error for a finalized design.

The relative error in body rotation
over a sweep of the tail due to this approximation is plotted in Fig.~\ref{fig:Tails_error_fig}a~as,
\begin{equation}
\label{eq:errorkin}
e_c(\xi_t, \eta) := \frac{\theta_{b,f}-\xi_t s_r}{\theta_{b,f}},
\end{equation}
where the exact final body orientation is found by integrating the connection \eqref{eq:A_nonlin} over the tail sweep;
an analytic expression for this function is derived in~\trsecref{tr:kinematics}.
For robots with $\eta\approx0$ or with $\xi_t\approx 0.5$, the error of this approximation is essentially negligible 
(less than $1\%$ for RHex or Tailbot).

\subsubsection{Tailed Body Dynamics}
\label{sec:taildynamics}

Defining for clarity the absolute tail angle, $\theta_t = \theta_b + \theta_r$,
and using the balance of angular momentum about the COM of each body,
the equations of motion for the full nonlinear tailed system are (see~\trsecref{tr:EOM}),
\begin{align}
\label{eq:NL_EOM}
\mathbf{M}(\theta_r)
\biggl[
{\begin{array}{cc}
\ddot{\theta}_b  \\
\ddot{\theta}_t \\
 \end{array} }
\biggr]
+
\biggl[
{\begin{array}{cc}
m_r l_b l_t \sin{\theta_r}\dot{\theta}_t^2  \\
-m_r l_b l_t \sin{\theta_r}\dot{\theta}_b^2
 \end{array} }
\biggr]
=
\biggl[
{\begin{array}{cc}
1  \\
-1 \\
 \end{array} }
\biggr]
\tau,
\end{align}
with an inertia tensor,
\begin{align}
\label{eq:NL_M}
\mathbf{M}(\theta_r) &= 
\biggl[
{\begin{array}{cc}
I_b+m_r l_b^2 &  -m_r l_b l_t \cos{\theta_r}  \\
-m_r l_b l_t \cos{\theta_r}  &I_t+m_r l_t^2 \\
 \end{array} }
\biggr]\\
 &=(I_b+ m_r l_b^2) 
\frac{\xi_t}{1-\xi_t} 
\biggl[
{\begin{array}{cc}
\frac{1-\xi_t}{\xi_t} &  -\eta \cos{\theta_r}  \\
-\eta \cos{\theta_r}  & 1 \\
 \end{array} }
\biggr].\notag
\end{align}

Inverting the inertia tensor yields an expression of the tailed body dynamics that, unlike the template~\eqref{eq:kappatau},
is both nonlinear and state-dependent,
\begin{equation}
\ddot{\theta}_b = \frac{\tau}{I_{d,t}(\theta_r)}-C_o(\theta_r,\dot{\theta}_b,\dot{\theta}_t),
\end{equation}
where $C_o$ is the Coriolis acceleration, and 
we define (by analogy to the template's driven inertia), the configuration-dependent inertia, $I_{d,t}$, 
\begin{align}
I_{d,t}(\theta_r) & = \frac{(I_t+m_r l_t^2)(I_b+m_r l_b^2)-(m_r l_b l_t \cos{\theta_r})^2}{I_t+m_r l_t^2 - m_r l_b l_t \cos{\theta_r}}\\
& = (I_b+ m_r l_b^2)\frac{1-\frac{\eta^2\xi_t}{1-\xi_t} \cos^2{\theta_r}}{1-\eta\cos{\theta_r}} . \label{eq:Idt}
\end{align}

In general, the Coriolis terms are negligible for tailed systems with small $\eta$, and
the anchoring can be accomplished with a constant (average) approximation of the driven inertia.
In the simplest case of body-centered 
tails (i.e.\ $l_b=0$, $\eta=0$), $I_{d,t}$ reduces to $I_b$ exactly and the Coriolis terms drop out,
allowing the choice of ${\Xi_{t,I_d}}(\mathbf{p_t}):=I_b$.

Most of the tails considered in Table~\ref{tab:properties} have $\xi_t\approx 0.5$, and so for these
tails consider $I_{d,0.5}:=I_{d,t}|_{\xi=0.5}$, which reduces exactly~to,
\begin{equation}
I_{d,0.5}(\theta_r) = (I_b+ m_r l_b^2)(1+\eta \cos{\theta_r})  . \label{eq:Idt_a}
\end{equation}
Integrating this function over a half tail sweep, $\theta_r\in [90^\circ ~ 270^\circ]$ (approximating the range of motion of many tails in Table~\ref{tab:properties}),
yields the best approximation for these bodies,
\begin{equation}
\bar{I}_{d,0.5}^* := (I_b+ m_r l_b^2)(1-\frac{2\eta}{\pi}).\label{eq:idave}
\end{equation}
In this paper we choose this as our morphological reduction for the driven inertia, i.e.,
${\Xi_{t,I_d}}(\mathbf{p_t}) := \bar{I}_{d,0.5}^*$,
although other choices may work better for some systems.
For each of the tailed systems surveyed in this paper the average deviation from~\eqref{eq:Idt} is less than 
$15\%$ over their actual tail range of motion;
for RHex the error is less than $2.5\%$.

\subsubsection{Final error due to approximate morphological reduction}
Using the template relations to constrain the power required to meet the righting task is subject to error from three sources: variation in the connection vector field, the changing inertia tensor, and the Coriolis accelerations.
This total error can be quantified over the tail design space ($\xi_t$, $\eta$) for a particular body/tail rotation task
by applying the nondimensionalization,~\eqref{eq:spatiotemporal}, to the nonlinear dynamics\footnote{This step isolates the effect of tail-specific geometry ($\xi_t$, $\eta$) from the remaining parameters, so that error can be 
quantified with respect to tail parameters alone.} 
and numerically integrating the resulting system 
(derived in~\trsecref{tr:ND_dyn}), with the optimal values of no-load speed and switching time from the template,
until the body comes to rest at a time $\tilde{t}_n$.
Defining the final body~error,
\begin{equation}
\label{eq:err_body}
e_b(\xi_t,\eta):= \tilde{\theta}(\tilde{t}_n) - 1
\end{equation}
and final time error,
\begin{equation}
\label{eq:err_time}
e_t(\xi_t,\eta):= \frac{\tilde{t}_n - \tilde{t}_f^*}{\tilde{t}_f^*},
\end{equation}
which are plotted in Fig.~\ref{fig:Tails_error_fig}b--c for a half sweep of the tail centered around $\theta_r = 180^\circ$.
Final error for this maneuver is less than $10\%$ across the large swath of parameter space containing the
examples found in the literature thus far; in particular, time and angle error fall within $2\%$ for RHex 
and within $4\%$ for Tailbot.

\subsection{Wheeled Morphological Reduction}
\label{sec:wheels}

A reaction wheel is a radially symmetric 
inertial appendage with mass centered at its joint, and
can be seen as a special case of a tail, with $l_t=0$; the appendage is simply a rigid body with 
inertia $I_w$ mounted a distance $l_b$ from the body's COM (Fig.~\ref{fig:TnA}).  
The parameter set for a reaction wheeled body is,
\begin{align}
\mathbf{p_w}:=[m_b, I_b, l_b, m_w, I_w , s_r, P, \omega_m, t_s, \theta_{b,f}, t_f], \label{eq:p_w}
\end{align}
where in general the wheel stroke, $s_r$, is infinite.

The connection field (and thus equivalent tail effectiveness) follows from (\ref{eq:A_nonlin}),
\begin{equation}
\frac{\partial \theta_b}{\partial \theta_r}  =-\frac{I_w}{I_w+I_b + m_r l_b^2} := -\xi_w. \label{eq:xiw}
\end{equation}
Here the vector field is a configuration-independent constant, as in the template,
and so the anchoring is exact.
The dynamics are found simply by setting $l_t=0$ in~\eqref{eq:NL_EOM} and~\eqref{eq:NL_M}; the 
nonlinear terms disappear and the dynamics become linear with driven inertia,
\begin{equation}
\label{eq:Idw}
I_{d,w} = I_b + m_r l_b^2.
\end{equation}
The non-identity components of the morphological reduction are thus chosen from~\eqref{eq:xiw} and~\eqref{eq:Idw}, as listed in Table~\ref{tab:morpho}.

\subsection{Limbed Morphological Reduction}
\label{sec:kinelimbs}

Unlike the tail and reaction
wheel anchors, whose kinematics' were more complex than the template's and consisted of a greater number of physical
parameters but still represented a single degree of freedom, an anchor model of a collection of limbs is truly a higher DOF mechanism.
The general problem of finding ``gaits'' in this larger shape space that extremize body rotation has been explored in \cite{hatton2011geometric};
here we consider the simpler cases that arise when the limbs are coordinated such that the effective shape space is one dimensional.
The resulting kinematics lie on a submanifold of the configuration space and, as we show,
are equivalent to the kinematics of the simpler template model. 
Hence the ``anchoring'' is accomplished through the active, closed loop control that coordinates the limbs.

In general, the effectiveness of an assemblage of limbs varies over their configuration space, even when coordinated.  However, two interesting
cases arise under certain conditions when all appendages are actively controlled to be parallel, that is each leg's relative angle is 
commanded to be either $\theta_i = \theta_r$ or $\theta_i=\theta_r+180^\circ$, for some common $\theta_r$.
Given $N$ limbs arranged with pivots 
in a line coincident with the body's COM (typically the centerline of the robot's body), a sufficient condition\footnote{The necessary condition is considerably more general, see~\trsecref{tr:limbs} for details. }
for configuration-independence of the connection field is that the limbs
are identical (each with mass $m_t$, length $l_t$, and MOI $I_t$), and that the pivot locations are symmetric across the body COM 
(as with the limbs of RHex, for example). 
Let $m_{tot}:=m_b+N m_t$ represent the total system mass, and $\ell_i$ 
the distance from body COM to the $i$th pivot location (generalizing the tail anchor's pivot offset $l_b$).
The expression of the total angular momentum  (derived in~\trsecref{tr:limbs}), 
reduces considerably in two illuminating examples, depending on the phasing of the limbs
(represented here by $s_i = \pm 1$, with $s$ negative for legs out of phase with $\theta_{r}$ by $180^\circ)$.
The full parameter set for an N-limbed system with the symmetry condition above is,
\begin{align}
\mathbf{p_\ell}:=[m_b, I_b, \ell_1, s_1, ..., \ell_N, &s_N, l_t,  m_t, I_t ,  \label{eq:p_ell} \\
& s_r, P,\omega_m, t_s, \theta_{b,f}, t_f], \nonumber
\end{align}
where here we assume for simplicity the limbs share the same range of motion $s_r$, and the power $P$ 
is taken to be the sum across all limbs. 

RHex has six identical legs arranged in symmetric pairs of pivots along the centerline of the body; that is, $N=6$, $\ell_1=-\ell_3$, 
$\ell_2=0$, and all legs have equal mass $m_t$ and length~$l_t$.  The pairs of legs are driven in anti-phase 
to generate an alternating tripod gait when walking or running, a condition that could be modeled 
here by taking $s_i$ negative for odd~$i$ 
and positive otherwise, so that $\sum_{i=1}^6 s_i = 0$.  
In the anti-phase case, the angular momentum reduces to, 
\begin{align}
H_{O,l} =& (I_p + N(I_t + m_t l_t^2))  \dot{\theta}_b +  N (I_t + m_t l_t^2) \dot{\theta}_r, \label{eq:HOlout}
\end{align}
where $I_p = I_b + m_t \sum\limits_{i=1}^N \ell_i^2$.  When all legs are in phase, $\sum_{i=1}^6 s_i = N$ and the angular momentum is,
\begin{align}
H_{O,l} =& (I_p + N(I_t + m_{rt} l_t^2))  \dot{\theta}_b + N(I_t + m_{rt} l_t^2) \dot{\theta}_r, \label{eq:HOlin}
\end{align}
with the subtle difference being the adjusted mass $m_{rt}:=m_b m_t /m_{tot}$, a generalization of $m_r$.
In either case, the connection field 
is constant, and thus the equivalent template effectiveness is error-free,
\begin{equation}
{\Xi_{\ell,\xi}}(\mathbf{p_\ell}) :=\xi_{\ell} = \frac{N (I_t + m_k l_t^2) }{I_b  + m_t \sum\limits_{i=1}^N \ell_i^2+N (I_t + m_k l_t^2)},
\label{eq:xileg}
\end{equation}
where $m_k = m_t$ when leg pairs are out of phase, and $m_k=m_{rt}$ when legs are in phase.  Since $m_t>m_{rt}$, anti-phase leg swings are more
effective than in-phase swings, as explored further in Section~\ref{sec:morphology}.

The multi-body dynamics of a robot with several phased appendages are considerably more complex than the developments
of the previous sections, and should be derived carefully for any particular case of interest.  Here we merely suggest 
a na\"{i}ve mapping based on the rotating inertia as expressed in the symmetric cases outlined in Section \ref{sec:kinelimbs}:
\begin{equation}
{\Xi_{\ell,I_d}}(\mathbf{p_\ell}) : =I_{d,\ell} = I_b  + m_t \sum\limits_{i=1}^N \ell_i^2, \label{eq:Idl}
\end{equation}
mapping the total input power across all limbs to the template.

\section{Comparative morphology and Scaling}
\label{sec:discussion}

Each of the diverse IR bodies of the previous section
can accomplish the given task, raising the question of how morphology shapes the available design choices.  The differences can be expressed 
and compared directly through each system's morphological reduction, as summarized in Table \ref{tab:morpho}. 
In this section we
examine the consequences of those anchoring relations and explore the implications for inertial 
reorientation at sizes large and small.

\subsection{Examples from the literature}
\label{sec:examples}
To facilitate our comparative approach, we present examples of IR machines from the literature in Tables~\ref{tab:properties_limb} \&~\ref{tab:properties} 
(compiled using the references shown and personal communications\footnote{Values differing from those in the cited references are more up-to-date or accurate.}).
As an interesting contrast to the mobile robots that are the focus of this paper, we included another notable example of terrestrial dynamic IR -- a small, off-road motorcycle (``dirt bike''),
as skilled riders are known to modulate the acceleration of the rear wheel to control orientation during leaps and tricks~\cite{brearley1981motor}.
Most use morphology designed specially for IR, but three machines (the two legged examples, and the motorcycle)
feature appendages designed for terrestrial locomotion that can be co-opted for aerial IR.
The mass range covered by the examples is surprisingly large -- over 300 fold among the tailed robots, and over three orders of magnitude in all.
This is not a comprehensive list of all robots harnessing inertial forces; notably, we have omitted devices where the tail moves in a plane far from the body COM,
as in~\cite{briggs2012tails}. However, the diversity of the chosen machines provides both a verification of the efficacy of the templates and anchors design
approach (noting the low final error for all machines) and enables some useful comparisons, as discussed in the following subsections.

\subsection{Selection of morphology for inertial reorientation}
\label{sec:morphology}

When is it appropriate to add a new appendage to a limbed body; and when is it better to assign the inertial appendage role to a tail rather than a reaction wheel?
In short, tails provide the most reasonable path to high values of effectiveness ($\xi \approx 0.5$ or higher), and are thus well suited to aggressive, dynamic maneuvers, while reaction wheels
provide infinite stroke over longer time scales.  Limbs may provide a middle ground, varying considerably in morphology across extant robots, and thus in effectiveness, and may provide some
IR capability without any additional payload.  

\subsubsection{Wheeled vs. Tailed Bodies}
The symmetric mass of a reaction wheel 
provides the advantage of simple, linear dynamics and infinite range of motion.
Of course, large wheels become 
cumbersome more quickly than a tail -- a practical reaction wheel could be no larger in diameter than a robot body's smallest dimension.
In natural systems, tails greater than body length are common, and thus we can expect larger effectiveness from tails than from
reaction wheels.  For example, between the comparably-sized Hexbug~\cite{casarez2013using} and TaYLRoACH~\cite{kohut2013precise}
(the former employing a pivot-centered double tail mass which acts like a wheel, and the latter an offset tail), the tailed design achieves roughly $15\%$ higher
effectiveness (0.44 vs 0.38) with $20\%$ lower appendage mass ($4g$ vs $5g$, Tables~\ref{tab:properties} \&~\ref{tab:properties_limb}).

Since wheels and limbs
need not incur the constrained range of motion 
suffered by practical 1-DOF tails,\footnote{More complex tails can escape this limitation in some maneuvers, e.g. the conical
tail motion generating roll in the falling gecko~\cite{jusufi2008active}.}
their effectiveness seems less important (i.e.\ it does not intrinsically limit body 
rotation) --  so why bother with a relatively bulky tail?
The answer is revealed though the power equation,~\eqref{eq:poweropt},
and its inverse dependence on tail effectiveness. For a given task, a doubling 
of $\xi$ reduces the power requirement by half.  Herein lies the fundamental limitation of low-effectiveness devices for fast reorientation:
a small  flywheel will require much more power than a relatively long tail for the same maneuver.  
The short time scales available for aerial reorientation in terrestrial robots suggest a limited role for internal reaction wheels,
but when this constraint is lifted (e.g.\ in space robotics\cite{umetani1989resolved}), such devices should be ideal.  
The motorcycle example in Table~\ref{tab:properties_limb} provides an instructive exception -- its IR ``appendage'' is driven by the machine's 
locomotive powertrain, resulting in the largest body mass-specific power (over $300~W/kg$) of any example here, enabling impressive aerial maneuverability in the right hands.
When retrofitting an IR appendage to an existing machine, the lower power requirements of a tailed design should lead to generally lower added mass than a less effective wheel.
For tails and wheels of comparable length scale, the advantage goes to the  wheel due to the subtle effect of the reduced mass
-- the offset tail pulls the system COM towards the tail as appendage mass increases, thus decreasing effectiveness
($m_r$ in $\xi_t$ is strictly smaller than~$m_t$).

\subsubsection{Limbed vs. Tailed Bodies}

For a given total added mass, a single appendage (tail) will generally provide larger effectiveness than two or more appendages.
The squared dependence of effectiveness on length makes elongate appendages most attractive; hence, dividing a 
tail into two limbs each with half the length and mass of the original appendage would entail a significant loss of performance (a pair of symmetric 
flywheels sees a similar disadvantage).
On the other hand, in many cases
(for example RHex), limbs also provide infinite stroke, can exceed body dimensions without negative consequences (unlike a reaction wheel), and will
by definition be already present on a legged terrestrial robot, eliminating any added cost or complexity.  
Machines with relatively long limbs will likely benefit most from this strategy (the quadruped Cheetah Cub achieves almost three times the IR effectiveness
of RHex with a third fewer limbs, see Table~\ref{tab:properties_limb}).
However, the use of these appendages for aerial reorientation may pose significant drawbacks, most notably a constraint on their 
final orientation upon landing (touching down feet-first is typically desirable).  
Explicit design for reorientation will likely also conflict with
other limb design priorities (for example, distal mass is typically a disadvantage when interacting impulsively with a substrate or when retracting the limb during the 
swing phase~\cite{seok2013design}). 
Still, in many cases even a limb designed for running may result in enough inertial effectiveness to be useful in small (but significant) rotations.
We will test this hypothesis in Section~\ref{sec:experiments}.

\subsubsection{Core vs. Appendage Actuation}

A tailed body and an actuated spine \cite{paper:mather-iros-2009, paper:xrhex_canid_spie_2012} can both
be represented by the same anchor model, but represent very different 
design propositions.
The primary advantage of a spine is that it may preserve the overall morphology 
(in particular volume and body envelope) by essentially separating the body
into two chunks with much lower MOIs (with $\xi_t \approx 0.5$ if the segments are similar).
Meanwhile, an added tail will in general extend the body envelope.
The major drawback of body-bending (as with using limbs for inertial reorientation) is that the final orientation of both segments is 
important if the legs of the robot are to hit the ground simultaneously \cite{paper:mather-iros-2009} -- as we show in Section~\ref{sec:experiments}, increasing the number of 
contact limbs when landing can greatly increase survivability.
Furthermore, existing robotic platforms (like RHex) cannot be substantially altered without a major redesign,
but their distal appendages may be relatively easy to add, subtract, or change.
The core actuation approach may have increased advantages outside the planar scope of this paper;
compare for example roll maneuvers in the falling cat~\cite{kane1969dynamical} against those of the falling gecko~\cite{jusufi_2010}.

\subsubsection{Maximizing tail performance}
Intuitively, tail effectiveness increases with tail mass, length and inertia, and decreases with the corresponding body parameters.
Minimizing tail offset (placing the joint close to the body COM) has the dual benefits of 
increasing performance and reducing nonlinearity (the MSU jumping robot~\cite{zhao2015controlling} comes closest to this ideal, while Tailbot could increase effectiveness by $10\%$ by centering its tail at the body's COM). Concentrating tail mass at the appendage's extreme produces
the most effectiveness per unit tail length (recall $l_t$ is the distance from pivot to tail COM, which if $I_t \neq 0$
is strictly less than the total tail length), and thus an idealized tailed body consists of a point-mass tail pinned at the body's COM.  Less intuitive is the trade-off between tail mass and length; clearly a given effectiveness can be 
accomplished with any number of combinations of each, though increasing tail mass eventually sees diminishing returns due to the effect of the reduced mass~\eqref{eq:mrdef}. By contrast, increasing tail length quadratically increases effectiveness.
RHex's relatively long tail achieves $75\%$ higher effectiveness than that of the Kangaroo robot with approximately the same fraction of overall mass dedicated to appendage.
At what point a tail's length becomes cumbersome is surely dependent on the constraints of other tasks and varies widely between applications,
but the examples of Table~\ref{tab:properties_limb} see tail lengths commonly exceeding one body length.

\subsection{Scaling of Inertial Reorientation}
\label{sec:scaling}

Agile mobile robots span an increasingly large size range, raising the question of whether inertial reorientation remains a practicable strategy for robots large and small.
In the next section, we design a tail for RHex with a task specification based on the righting performance of Tailbot, a robot approximately one fiftieth of RHex's mass.
How will this mass difference dictate changes in morphology or mass-specific motor power?
Because $\xi$ is dimensionless and dependent only on morphology, isometrically~\cite{blackburn_1994_animal} 
scaled robots are kinematically similar -- for a given appendage rotation, the body rotation will be identical at any size scale.  
However, the power required for a maneuver will vary with size.

Consider a robot
isometrically scaled by a length $L$.   Assuming uniform density,  the
robot's  mass will scale by $L^3$ and its inertia by $L^5$.  
 If the robot were required to reorient through the same angle
in the same time regardless of size, then by substitution into~\eqref{eq:poweropt}, (replacing $I_d$ with $L^5$ and dividing both sides of the inequality by $L^3$)
we would require power per unit robot mass (power density of the whole machine) $P_d \propto L^{2}$.  However, because gravity is constant, $g$, a larger robot
will fall slower relative to its length (i.e.\ dynamic similarity~\cite{alexander1983dynamic}).  For a free
fall distance of $h\propto L$, the time available is $t_f = \sqrt{2h/g} \propto L^{1/2}$. 
Therefore, from \eqref{eq:poweropt}, the required power per unit robot mass,
\begin{equation}
\label{powerscale}
P_d\propto \frac{I_d}{m t_f^3} \propto \frac{L^5}{L^3L^{3/2}} = L^{1/2},
\end{equation}
scales as the square root of length.  This indicates that inertial reorientation gets mildly more expensive at large size scales; larger
robots may suffer reduced performance, or must dedicate a growing portion of total body mass to tail actuation (or, noting 
the inverse relationship with $\xi$, to increased tail effectiveness).  However,
RHex and Tailbot span a characteristic length range of almost four fold without dramatic differences in ability (see Fig.~\ref{fig:rhexfall});
in fact, the smaller machine dedicates more body mass to its tail motor than RHex ($6.9\%$ vs. $3.3\%$), even as
the larger machine has relatively higher body inertia (an isometrically-scaled Tailbot of RHex's mass would have $I_b = .11~kg m^2$, 
almost $30\%$ lower than RHex).
In this case, differences in actuator performance  trump scaling -- Tailbot uses a low-quality brushed motor, 
while RHex's higher quality components,~\cite{paper:xrhex_canid_spie_2012}, allow it to escape the penalty of size.

Intriguingly, 
generalization of the IR template dynamics
suggests that \eqref{powerscale} may govern scaling of other power-limited self-manipulation tasks, including 
aspects of legged locomotion.
Consider a robot with its feet planted firmly on the ground, rotating its body in the yaw plane
about an actuated hip. This situation could be modeled by a single rigid body, connected to the ground by
a motor -- that is, the system can be modeled by the IR template, 
considering the ground to be the ``appendage'', with $I_a$ infinitely large and $\xi = 1$.
Power for reorientation for this grounded reorientation task scales as in~\eqref{powerscale}.\footnote{
The scaling of relevant time scale (during a single step) again follows dynamic similarity, as stride frequency in running
scales with $\sqrt{L}$~\cite{alexander1983dynamic}.}
In this simplified scenario, power-limited reorientation scales identically whether the body rotation is driven by inertial or ground reaction forces;
we therefore hypothesize that inertial appendages may enhance agility at any size scale permitting legged maneuverability.

\setlength{\tabcolsep}{0.43em}
\begin{table}
  \centering	    
    \caption{Comparison of physical properties for limbed or wheeled systems with the capability for aerial reorientation. Unlike the tailed examples, these machines anchor without error.}
	  \label{tab:properties_limb}
    \begin{tabular}{ l c c c c}
      \toprule
      \midrule
      Attribute& RHex & Cub & Hexbug& Dirt bike \\
      \midrule
      Citation & & \cite{paper:heim2015tail} & \cite{casarez2013using} & \cite{brearley1981motor}   \\
      Number in error figure & 12 & 11 & 9 & 10 \\ 
     Appendage Type & Limbs & Limbs & Wheel & Wheel  \\
      Body length (cm),$L$ & 57 & 21 & 5 & 140 \\
      Body mass (g), $m_b$ & 7500 & 1300 & 40 & $105\!\!\times\!\!10^{3}$ \\
      App. mass (g), $m_t$ & 63 & 52 & 5 & $10\!\!\times\!\!10^{3}$ \\ 
      App. offset (cm), $l_b, \ell_i$ & 25, 0, 25 & 10, 10 & 2.5 & 70 \\ 
      App. length (cm), $l_t$ & 10 & 6.3 & 0 & 0 \\ 
      Body inertia (kgm$^2$), $I_b$ & 0.15 & $9.8\!\!\times\!\!10^{\minus3}$ & $17\!\!\times\!\!10^{\minus6}$ & 20 \\ 
      App. inertia (kgm$^2$), $I_t$ & $0.46\!\!\times\!\!10^{\minus3}$ & $0.14\!\!\times\!\!10^{\minus3}$ & $12\!\!\times\!\!10^{\minus6}$ & 0.4 \\ 
      \midrule
      Effectiveness, $\xi_{\ell}, \xi_w$ & 0.037 & 0.096 & 0.38 & 0.016  \\ 
      Driven inertia (kgm$^2$), $I_d\!\!\!$ & 0.17 & 0.012 & $19\!\!\times\!\!10^{\minus6}$ & 24  \\ 
      Peak motor power (W) & 2052 & 23.3 & 0.34 & $33\!\!\times\!\!10^{3}$  \\ 
      Range of motion, $s_r$ & $360^\circ$ & $180^\circ$ & $360^\circ$ & $360^\circ$  \\ 
      App. speed (RPM), $\omega_m$ & 434 & 77 & 916 & 1200 \\
      \bottomrule\\	      
    \end{tabular}
    \vspace{-1ex}
\end{table}

\begin{table*}[tb]
  \centering
    \caption{Comparison of physical properties for tailed systems with the capability for aerial reorientation.
    }
    \label{tab:properties}
    \begin{tabular}{ l c c c c c c c c c c c}
      \toprule
	 \midrule
        Attribute& RHex & Tailbot& TaYLRoACH& 2DOF Tailbot& Jumper& Kangaroo& Jerboa& Cub\\
        \midrule
        Citation & & \cite{chang-siu-iros-2011} & \cite{kohut2013precise} &   \cite{2DOFtailbot} & \cite{zhao2015controlling} & \cite{liu2014bio} &  \cite{CompositionHoppingTR} &  \cite{paper:heim2015tail}\\		
        Number in error figure & 1 & 2 & 3 & 4 & 5 & 6 & 7 & 8 \\ 
        Body length (cm), $L$ & 57 & 11.7 & 10 & 13.5 & 7.5 & 46 & 21 & 21 \\ 
        Body Mass (g), $m_b$ & 8100 & 160 & 46 & 105 & 25.1 & 5030 & 2270 & 1250 \\ 
        Tail Mass (g), $m_t$ & 600 & 17 & 4 & 70 & 1.4 & 371 & 150 & 310 \\ 
        Tail offset (cm), $l_b$ & 8 & 4.5 & 5 & 5.2 & 1 & 15.6 & 3 & 10 \\ 
        Tail length (cm), $l_t$ & 59 & 10.3 & 10.2 & 7.3 & 6.8 & 17.7 & 30 & 16.8 \\ 
        Body Inertia (kgm$^2$), $I_b$ & 0.15 & $154\!\!\times\!\!10^{\minus6}$ & $39.6\!\!\times\!\!10^{\minus6}$ & $210\!\!\times\!\!10^{\minus6}$ & $9.3\!\!\times\!\!10^{\minus6}$ & 0.05 & 0.025 & 0.01 \\ 
        Tail Inertia (kgm$^2$), $I_t$ & 0\textsuperscript{\ref{fn:Itzero}} & 0\textsuperscript{\ref{fn:Itzero}} & 0\textsuperscript{\ref{fn:Itzero}} & $479\!\!\times\!\!10^{\minus6}$ & $6.4\!\!\times\!\!10^{\minus6}$ & 0.0172 & 0\textsuperscript{\ref{fn:Itzero}} & $875\!\!\times\!\!10^{\minus6}$ \\ 
        \midrule
        Nonlinearity, $\eta$ & 0.136 & 0.437 & 0.49 & 0.227 & 0.072 & 0.339 & 0.1 & 0.529 \\ 
        Tail effectiveness, $\xi_t$ & 0.5587 & 0.4683 & 0.4396 & 0.6848 & 0.5705 & 0.3235 & 0.3351 & 0.3911 \\ 
        Peak Motor Power (W) & 342 & 4 & 2.5 & 1.75 & 0.257 & 19 & 426 & 5.82 \\ 
        Driven inertia (kgm$^2$), ${I}_d$ & 0.141 & $145\!\!\times\!\!10^{\minus6}$ & $37.2\!\!\times\!\!10^{\minus6}$ & $283\!\!\times\!\!10^{\minus6}$ & $9.02\!\!\times\!\!10^{\minus6}$ & 0.0482 & 0.0236 & 0.0092 \\ 
        Range of motion, $s_r$ & $172.5^\circ$ & $255^\circ$ & $265^\circ$ & $135^\circ$ & $280^\circ$ & $220^\circ$ & $180^\circ$ & $110^\circ$ \\ 
        Tail speed (RPM), $\omega_m$ & 356  & 3000 & 400 & 320 & 1000 & 240& 353 & 77 \\
        \midrule
        Error, final angle~\eqref{eq:err_body} & $-1.29\% $ & $ -1.90\% $ & $ -1.26\% $ & $ -4.78\% $ & $ -0.630\% $ & $ 3.48\% $ & $ 1.48 \% $ & $ 0.507 \%$ \\ 
        Error, final time~\eqref{eq:err_time} & $1.20\% $ & $ 3.92 \% $ & $ 4.91 \% $ & $ 0.105 \% $ & $ 0.836 \% $ & $ 5.75 \% $ & $ 1.94\% $ & $ 6.59\% $  \\ 
	      \bottomrule\\	      
	      \end{tabular}
      \vspace{-5ex}
\end{table*}

\section{Design for Inertial Reorientation}
\label{sec:motorsizing}

In this section, we present examples of the complementary design problems of \emph{Body Selection} and \emph{Performance Evaluation} (introduced in Section~\ref{sec:kinematics})
by exploring IR morphology for RHex.
The first step in the \emph{Body Selection} problem, \ref{problem:body},
is to specify the task or set of tasks required of the machine (i.e.\ parametrizing~\eqref{eq:Sr}); the task and other (external) concerns will determine the 
overall morphology, subject to the trade-offs discussed in the previous section. With a body plan chosen, 
the designer is then free to pick any set of physical parameters in $\mathcal{R}_i$
that best meets performance needs outside the reorientation task. 
A na\"{i}vely  rational  design approach might introduce a cost function, $C(\mathbf{p}_i)$, expressing the impact of the IR morphology on some
other critical task (e.g.\ legged locomotion) or penalty (e.g.\ parts cost) and solve the resulting constrained minimization task. However, it is notoriously
difficult to encode robustness within the rigid optimization framework. Robots, putatively general purpose machines, will typically
be assigned multiple critical tasks, oft-times with conflicting objectives (e.g.\ fast locomotion and steady perception). 
More frequently, legacy constraints imposed by a robot's existing design will further reduce the design 
problem to the selection of one or two parameters, precluding the possibility of an optimized design.  
Every design problem (whether of tails, limbs, flywheels or other morphology)
will likely entail its own set of constraints, assumptions and objectives which must be chosen such that \eqref{eq:pullback}
results in a suitable and unique design solution.

\begin{figure}
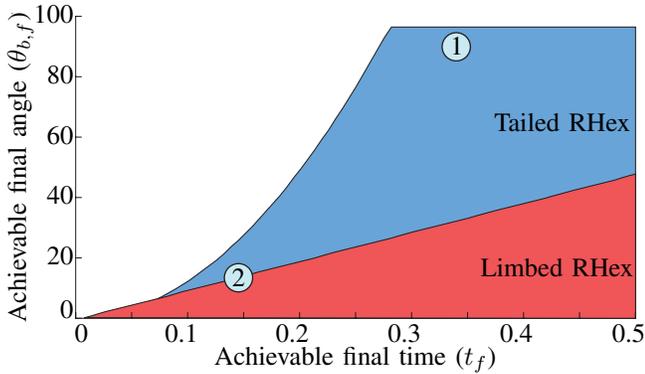

    \centering
    \def\svgwidth{8.5cm}
    \include{taskspace}
    \caption{The regions of task space (a projection onto the $\theta_{b,f}$ and $t_f$ components of the feasible set $\mathcal{R}$) accessible by 
    two instantiations of IR morphology on RHex for the reorientation task,~\eqref{eq:Sr}. The tail is limited by power for the quickest tasks,
    and by stroke for slower maneuvers; its higher effectiveness allows far more useful rotation
    at relevant time scales. The numbers indicate the two experimentally-validated tasks: 1) tailed reorientation in one body-length
    fall and 2) limbed reorientation during a leap. Both tasks fall within the tailed body's feasible set, but task (1) exceeds 
    the limbed body's capability.   }
    \label{fig:taskspace}
\end{figure}

In the \emph{Performance Evaluation} problem, \ref{problem:performance},
the fixed design restricts the system performance to a subset of task space (the projection of 
the feasible design set $\mathcal{R}$ onto the $(\theta_{b,f},\,t_f)$ subspace).
This region can be computed for set values of $\xi$, $I_d$, $P$, $s_r$, and $\omega_m$ by
using~\eqref{eq:TemplateEvalConstraints} to query the feasibility of a task
(values of $t_f$ and $\theta_{b,f}$), selecting the switching time $t_s$ to satisfy the final angle condition, if possible.
A fixed template design will necessarily be suboptimally geared for most tasks in the feasible task subspace;
the cost of this suboptimality (along with that of submaximal current limit) can be calculated through
the changing power cost, $k_p$, in \eqref{eq:poweropt}.
We compare the achievable task subspace for two implementations of IR morphology on RHex in Fig.~\ref{fig:taskspace}, and list values of $k_p$ where applicable.

In practice, the design process will use both the selection and evaluation problems to settle on a solution both practicable and task-feasible. 
Starting with the \emph{Body Selection} problem (parametrizing a task and choosing a body plan), the designer should first use 
$\mathcal{R}_i^*$ to achieve a rough design, as the reduced (gearing-optimal) space and simpler form of the constraints will highlight the 
consequences of any choices (fixing legacy-constrained physical properties, or adding constraints to satisfy other task objectives). 
Since practical concerns will further limit parameter choices (e.g. the optimal powertrain
is not likely to exist as an off-the-shelf product), the designer should then use
\emph{Performance Evaluations} of several candidate designs to find a feasible and physically realizable design.
A major advantage of this approach over a straightforward optimization is that the effects of the inevitable deviations 
from optimality can be quantified and compared (e.g. through $k_p$), thus informing the designer's concessions to practicability.

Real-world actuator selection is constrained by factors beyond rated power, as used in the preceding sections.
Choosing a powertrain for a real system also involves characterizing motors by their 
electrical (current, voltage), thermal, legacy (constraints of the robot's body), 
physical (size, mass), financial, and labor costs, as we show in the selection of the final motor for 
the following design experiments.

\subsection{Appendage design for RHex}
\label{sec:xrlapp}

\subsubsection{Tail payload}
\label{sec:xrltail}

As an example of the \emph{Body Selection} problem, \ref{problem:body}, we designed a tail for RHex by first specifying the task parameters, and then 
using $\mathcal{R}_t^*$ to guide the selection of the remaining
values in $\mathbf{p}_t$; the robot's existing morphology further constrains our choices to a subset of $\mathcal{R}_t$. 

In the interest of direct comparison with Tailbot~\cite{chang-siu-iros-2011}, we selected task specifications based on 
replicating one element of the smaller robot's behavioral repertoire: a reorientation of $\theta_{b,f} = 90^\circ$ in the course of falling one body length, $L$.
For RHex, this translates to the task specification,
\begin{align}
\theta_{b,f} &= 90^\circ, \qquad
t_f = \sqrt{\frac{2L}{g}} \approx 0.34~s,	\label{eq:XRLtask}
\end{align}
where $g$ is the gravitational acceleration.
As discussed in Section~\ref{sec:morphology}, the large effectiveness easily achieved by a tail makes that 
morphology the most attractive choice for this relatively aggressive maneuvering task (significantly decreasing the 
actuator requirements through the power equation,~\eqref{eq:poweropt}).

Of the full set of tailed-body parameters $\mathbf{p_t}$,~\eqref{eq:p_t}, 
two (body mass and inertia) were already set by RHex's existing body morphology, and a third (pivot location) was constrained by the body's envelope.
Confident that we could make the tail very nearly a point mass on a near massless rod
(thus maximizing effectiveness per unit mass and length), we further eliminated $I_t$.\footnote{The mass-centered rotational inertia of a small mass on a light rod
is far smaller than the offset inertia, $m_r l_t^2$; the $I_t$ value of this tail was therefore reported as zero in the cited work. \label{fn:Itzero}}
While Tailbot was a special-built machine,
the tail for RHex was added to the existing platform as a modular payload~\cite{paper:xrhex_canid_spie_2012}
and as such the range of motion is significantly lower than Tailbot's.  
As the design of the modular payload system limits maximum tail sweep to $180^\circ$ regardless
of pivot position, we centered the tail along the body axis to minimize $l_b = 8 ~ cm$ (maximizing effectiveness, reducing $\eta$ and further motivating the efficacy of \eqref{eq:TemplateConstraints});
a small safety margin to avoid
collision with the body reduced stroke slightly further to $s_r \leq 172.5^\circ$.  
With the selection of this range of motion limit, tail effectiveness is constrained by~\eqref{eq:TemplateConstraints} to
$\xi_t \ge 0.522$, leaving the question of the balance between tail length and~mass.
The addition of weight to RHex
via external payload has known (small) performance costs, while the addition of a long tail has  unpredictable and potentially large
consequences on capability outside of aerial righting;   
we therefore chose to minimize tail length
by selecting an additional
mass constraint based on previous experiences with modular payloads,
$m_t \leq 0.6~kg$ (giving $m_r = 0.56~kg$).
With $I_t \approx 0$, the minimum tail length to meet the effectiveness requirement can be found directly from the definition of $\xi_t$ (see Table~\ref{tab:morpho}), and is
$l_t \geq 0.55~m$.
As assembled, RHex's actual tail effectiveness is slightly larger than required, and is about $20\%$ larger than that of Tailbot (see Table~\ref{tab:properties}), 
as needed to achieve feasibility respecting the stroke constraint consequent upon the roughly $30\%$ reduction of its tail stroke relative to that of the smaller machine.  

Meeting the body stroke specification fixed all parameters save motor power, 
which is constrained by the second inequality in~\eqref{eq:TemplateConstraints};
the smallest allowable value of $P$ satisfying this constraint
is approximately $39~W$, with an optimal no-load speed just over $2~Hz$.  
The Maxon pancake motors that drive RHex's legs are rated for $50~W$ continuous operation,
and can achieve transient output up to $342~W$ \cite{paper:xrhex_canid_spie_2012}, but practical concerns
including thermal safety limit current to $12~A$, just $33\%$ of transient stall current (see Appendix~\ref{app:cl}).  
A putative design using these motors falls well within $\mathcal{R}_t$ despite their suboptimal gearing of 28:1
(effective $\tilde{\omega}_m \approx 1.0$, $\beta = 0.33$ giving $k_{p,t} \approx 11$ for this task, roughly four times higher than optimal);
we found that mitigation of integration issues outweighed any possible weight savings that could be had by choosing a smaller motor with more optimal gearing.
The chosen design is capable of rotating the body to $90^\circ$ within a predicted final time of approximately $300~ ms$, 
well within the performance specification. This tailed design is tested in Section~\ref{sec:tailex}.

\subsubsection{Flailing limbs}
\label{sec:xrlflail}
A highly attractive alternative to the added complexity of a tail is to simply use RHex's existing limbs, preferably in the in-phase condition so as 
to land on all six simultaneously.  
The total reorientation effectiveness, as predicted by~\eqref{eq:xileg}, is $\xi_{\ell} = 0.037$ (see Table~\ref{tab:properties_limb}). 
With $\mathbf{p_\ell}$ fixed by the existing design, we can query~\eqref{eq:TemplateEvalConstraints} to check the feasibility of this body with respect to the task,~\eqref{eq:XRLtask}.
The unlimited limb rotation means the design trivially meets the stroke specification, but not within the final time (Fig.~\ref{fig:taskspace}).
The very low effectiveness of the combined limbs necessitates almost eight full swings of the limbs to complete the body stroke requirement of~\eqref{eq:XRLtask}, and  
thus a substantially different power train than is used for terrestrial locomotion: the optimal no-load
speed for the limbed design of 2,178 RPM is almost 13 times higher than RHex's maximum leg speed.

While RHex's existing morphology is inadequate for this highly agile task, its limbs still provide a potentially useful IR capability -- the limbed 
system can rotate $32.3^\circ$ in one body-length of fall, or over $50^\circ$ in the $1.36~m$ fall we used to test RHex's tail (see Fig.~\ref{fig:taskspace}).
Such small reorientations
could be significant especially when running, where the nominal body orientation varies a similarly small amount \cite{ws:johnson-gait-2014}.
One full rotation of RHex's six limbs produces $13.3^\circ$ of body rotation, and its powertrain can achieve this reorientation in as little as 
$150~ms$.  This new reorientation task fits easily into the aerial phase of a single leap, usefully allowing modulation of landing angle;
we test it empirically in Section~\ref{sec:limbex}.

\subsection{Experiments on RHex}
\label{sec:experiments}

\begin{figure}[t]
    \centering
    \includegraphics[scale=.25]{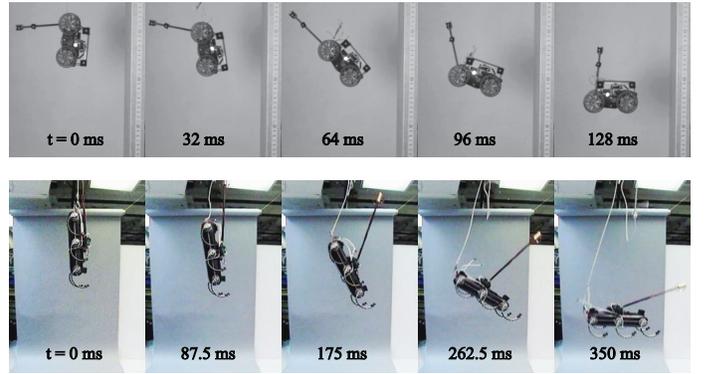}
    \caption{Dynamically similar aerial righting in two robots spanning a 60-fold mass range: Tailbot (top) and RHex (bottom). Each machine 
    		rotates $90^\circ$ in approximately one body length of fall.}
    \label{fig:rhexfall}
\end{figure}

\subsubsection{IR Task Implemented on the Tailed-Body RHex Design}
\label{sec:tailex}
\begin{figure}
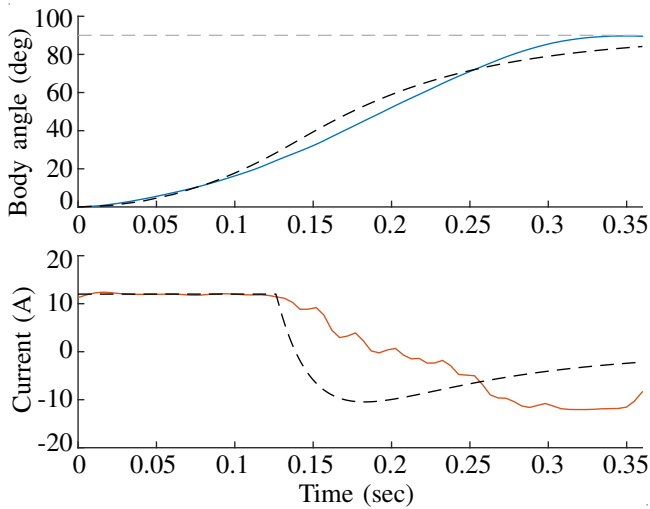

    \centering
    \def\svgwidth{8.5cm}
    \include{vertical_log}
    \caption{Logged data from a tailed robot experiment. (Top) Body angle, from high-speed video (blue) and predicted by template with PD controller (dashed); (Bottom) motor current, applied (red) and predicted by template (dashed).
 	       Disagreement between model and template is primarily due to unmodeled compliance in the tail pivot and shaft.  
     }
    \label{fig:vert_log}
\end{figure}

As an anecdotal validation of the foregoing scaling arguments, we
conducted a series of inertial reorientation experiments on RHex (Figs.~\ref{fig:rhexfall} \&~\ref{fig:vert_log}).
In the first experiment, the robot was dropped nose first from a height of $1.36~m$ (over $8$ times the standing
height and $2.7$ times the body length, though we still required the robot to meet the task specification in~\eqref{eq:XRLtask}).  

We implemented 
a PD (proportional-derivative) controller on the internal angle in lieu of a bang-bang controller -- 
as discussed in Appendix~\ref{app:altcont}, the saturated PD controller 
converges to the bang-bang design
given large enough gains. In practice, the sensor bandwidth and other unmodeled effects result in oscillation around the regulated angle, 
so we relaxed the high gain requirement slightly,
accepting the slight performance cost in favor of the extra robustness provided by the closed-loop design 
(the effective current switching time of approximately $0.22$ seconds corresponds to a $\tilde{t}_s \approx 2.85$, slightly later
than the optimal $2.40$ for the chosen gearing and current limit).
The legs were simultaneously controlled to point towards the ground.

Data from a typical reorientation experiment can be seen in Fig.~\ref{fig:vert_log}; we logged sensor data and shot high-speed video
at 210 frames per second.
The robot rotated to within $1^\circ$ of horizontal before landing on all six legs, taking about $350~ms$ to complete the reorientation,
corresponding to a dimensionless halting time of approximately $4.55$, and a corresponding power cost $k_p \approx 23.5$ --
more than nine times optimal, and twice the cost predicted in section~\ref{sec:xrltail}, due to the suboptimal controller and several unmodeled effects.
Drivetrain losses decreased output torque by roughly $25\%$ and the tail mount and carbon fiber shaft exhibited substantial elasticity, causing acceleration lag
and increasing the deviation from bang-bang torque application.
Despite the high cost of the suboptimal design, the robot completes the task within $3\%$ of the target time. As further verification of the template, we 
added the PD controller and drivetrain efficiency losses to the model; simulation of this more accurate (suboptimal) template is plotted against experimental data in Fig.~\ref{fig:vert_log}.

When dropped with the tail controller off, the robot impacted the ground nose first, with
only the front pair of legs in support. The impacted limbs quickly snapped, allowing the robot's body to strike the ground, 
causing internal damage. 
We therefore conclude that the active inertial tail substantially expands the task space of RHex by tripling the
number of support limbs available for impact mitigation (thereby roughly tripling the strain energy tolerable before failure and increasing
the survivable falling height). 

\begin{figure*}[t]
    \centering
    \includegraphics[scale=.4,trim = 0mm 0mm 0mm 10mm, clip]{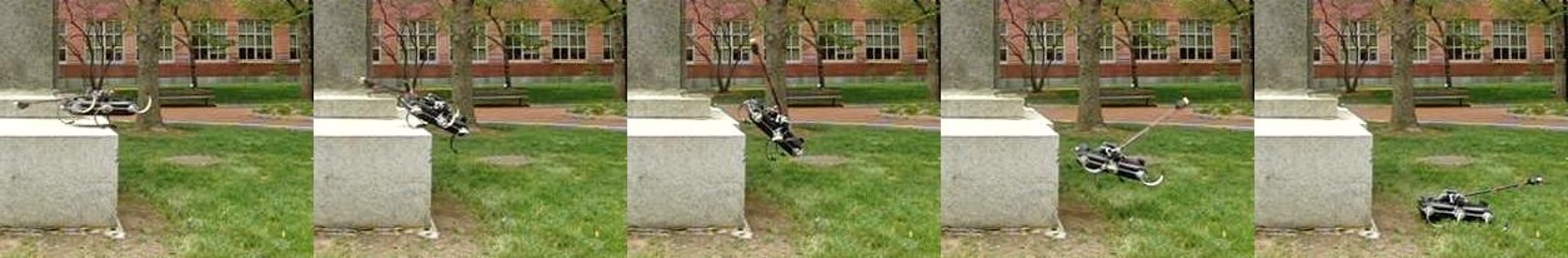}
    \caption{RHex surviving a run off a cliff outdoors.}
    \label{fig:rhexcliff}
\end{figure*}

To demonstrate this new ability for RHex in a practical task, the robot was also tested outdoors
running along and then off of a $62~cm$ (3.8 hip-heights or 1.2 body-lengths) cliff. 
This stabilization task is governed by a different set of performance constraints that could be probed analogously to our approach
to the reorientation task in Section~\ref{sec:reduced};
\footnote{The stabilization task could be specified by keeping the body angle within some allowable deviation over a time horizon $t_f$
by swinging the tail to mitigate an impulse characterized by $\tilde{H}$ as in~\eqref{eq:xidef}. Constraints analogous to~\eqref{eq:TemplateConstraints} could be 
derived by solving the template kinematics and dynamics subject to this new task.}
in lieu of a more exhaustive analytic exploration of this task space, we note
that this fall nearly saturated RHex's tail stroke and likely represents a near-limit for full stabilization at this running speed.
The robot's inertial sensors detect the cliff 
upon initial body pitch, then actuate the tail according to the same PD  
control policy, landing the robot on its feet (Fig.~\ref{fig:rhexcliff}). 
As with the indoor experiments, a test with RHex running off a cliff with a passive tail confirmed 
that the robot lands nose first.

\subsubsection{IR Template Anchored  on RHex as a Limbed Body}
\label{sec:limbex}

Finally, a third set of experiments tested the ability of RHex's legs to work as inertial tails. The existing 
limb design is incapable of achieving the original task ($90^\circ$ body rotation in $0.34~s$, Fig.~\ref{fig:taskspace}),
and so instead a feasible task consisting of a single rotation is used instead to test the flail kinematics. As first
reported in~\cite[Sec.~IV-C.5]{johnson_selfmanip_2013}, after leaping vertically into the air all of the legs
were recirculated together to the same landing angle. 
Using the legs in phase for this experiment allows the robot to land on all six, though using the legs 
out of phase would have increased the effectiveness by about $3.5\%$ (Section~\ref{sec:kinelimbs}).\footnote{As in the tailed trials, we
used a PD controller on the internal angle (see Appendix~\ref{app:altcont}), however here this single control effort was pulled back into the 
more complicated limbed body through an anchoring controller (specifically, six independent PD controllers each regulating a limb
to the common commanded position.} 
The limbs were rotated clockwise in the first experiment
and counterclockwise in the second for a net difference of $360^\circ$ in stroke; the difference in final body angle between the two cases was 
$14^\circ$.  
While the leap gave only enough time for a single revolution of the limbs, the resulting body rotation
made an appreciable difference in the quality of the landing, supporting our
hypothesis that IR with even unspecialized limbs can be useful.

\section{Conclusion}
\label{sec:Conclusion}

As mobile robots proliferate in the complexity of both their morphology and behavioral scope, there is a growing need for principled methodology relating their
body design to their capability. 
The templates-and-anchors approach adopted for this paper provides a unifying framework for 
the comparative morphology of robots (and even animals) and a practical approach to the design and evaluation of 
inertial reorientation performance on real robots. 
We defined the Inertial Reorientation template, the simplest model of an IR maneuver, equipped with a DC motor-like model
parametrized by peak output power.  
The template revealed the particular importance of a single parameter, defined here as IR effectiveness, which prescribed both the appendage rotation needed to move a body and 
the power needed to do so in fixed time.
Dimensional analysis of template behavior revealed that a relatively 
modest increase in power density (growing with the square root of length) should be required  to retain righting performance as platform size increases.  

 The model's linear dynamics, along with a bang-bang
controller, enabled analytic  solution of the template's single-switch reorientation behavior, revealing a simple relationship between morphology 
and performance, described by the task-feasible set $\mathcal{R}$,~\eqref{eq:TemplateEvalConstraints}. 
We then showed how the feasible set could be ``pulled back''
through a more complex body's \emph{morphological reduction} (defined as the possibly-approximate mapping between parameter spaces of real robot body and abstracted template),
to provide design restriction to a more usefully diverse set of machines.
The resulting set of feasible real designs~\eqref{eq:pullback} retains enough freedom (e.g. allowing length to trade for mass) to afford 
some ``optimization'' in the sense of minimizing the impact of the design on other task abilities. 
In practice, concessions to practicality will necessitate deviations
from optimality; fortunately, our framework gives the designer flexibility to compare candidate (suboptimal) designs and even quantify the performance cost of 
those compromises (e.g. through~\eqref{eq:kpdef}).

Our approach facilitated the design of a tail for RHex, enabling inertial reorientation capabilities dynamically similar to the much-smaller Tailbot.  A separate anchoring to 
the same template quantified the capability of RHex's existing appendages (its six semi-circular legs) to
produce useful reorientation in their own right, and revealed a preferred posture for doing so.  
A recent proliferation of tails (and other high-effectiveness appendages) for inertial righting allows us to calculate and compare effectiveness 
across a number of independent designs; generally their effectiveness is close to $0.5$, where the connection 
becomes configuration-independent.  As a result, these designs anchor nicely to the IR template with 
relatively low error. We expect that most well-designed appendages will fall within this paradigm.

The constraints making up $\mathcal{R}$ and its gearing-optimal refinement~\eqref{eq:TemplateConstraints} revealed general principles of design for
righting morphology, while the morphological reductions provided crucial insight into the tradeoffs of each body type; we provide a detailed discussion 
in Section~\ref{sec:discussion}.
Tails are a natural choice for fast, large amplitude inertial reorientation, owing to the ease at which they can be designed for high effectiveness values
without disrupting the existing platform morphology.  However, as legged robots increase the numbers of DOFs in limbs and body alike, these affordances should
 provide compelling sources of inertial reorientation as well.  
In practice, the choice of anchor morphology for enabling inertial reorientation in a robot is tightly coupled to overall function with
respect to its mission, historical and other constraints on body design, and the task-specific rewards for high reorientation performance.

While the present analysis focuses on purely aerial maneuvers, inertial appendages also show promise in a variety of terrestrial
tasks, stabilizing or actuating turns \cite{kohut2013precise,patel2013rapid}, or stabilizing pitch over obstacles \cite{chang-siu-iros-2011}.
Likewise, inertial appendages have utility beyond the sagittal plane for aerial maneuvers, with out-of-plane appendage swings capable of effecting body rotations 
in yaw and roll as well as pitch, e.g.\ \cite{jusufi_2010,2DOFtailbot}.
We postulate that tail effectiveness will remain a useful metric in these arenas as well, though the
analysis of the dynamics and control affordance underpinning such behaviors remains an open problem.

\section*{Acknowledgments}

Some of the data in Tables~\ref{tab:properties} \&~\ref{tab:properties_limb} was collected through 
personal communication with the authors of the cited work, and we would like to thank  
Nick Kohut, Jianguo Zhao, Pei-Chun Lin, Avik De, and Steve Heim for contributing this morphometric data.
The motorcycle example used information from Zero Motorcycles, Zero FX.
The authors would like to thank Praveer Nidamaluri for building the X-RHex tail,
Aaron Effron and the four anonymous reviewers for their notes on the manuscript,
as well as David Hallac, Justin Starr, Avik De, Ryan Knopf, Mike Choi, Joseph Coto, and Adam Farabaugh for
help with RHex and experiments.

\bibliographystyle{IEEEtran}
\bibliography{IEEEabrv,tails}
\appendix

\subsection{Generalized Template-Anchor Relationship}
\label{app:anchors}
This section develops a general framework for the anchoring of more complex dynamical systems to a simpler
template dynamical system~\cite{Full-JEB-1999}. This is a more general notion of anchoring than in, 
e.g.,~\cite{altendorfer2004stability}, which requires the template dynamics be, ``conjugate to the
restriction dynamics of the anchor on an attracting invariant submanifold,'' or in~\cite[Sec.~1.2]{Altendorfer-AR2001},
which seeks, ``controllers whose closed loops result in a low dimensional attracting
invariant submanifold on which the restriction dynamics is a copy of the template.''

In particular consider two dynamical systems: the ``template'', $X$, and the ``anchor'', $Y$. Each system
has a state ($x \in \mathcal{X}$ and $y \in \mathcal{Y}$, respectively), 
control input ($u_X \in \mathcal{U}_X$ and $u_Y \in \mathcal{U}_Y$, respectively),
parameter set ($p_X \in \mathcal{P}_X$ and $p_Y \in \mathcal{P}_Y$, respectively),
and dynamics ($\dot{x}=f_X(x,u_X,p_X)$ and $\dot{y}=f_Y(y,u_Y,p_Y)$, respectively).
The template is the simpler model, so in general, $\dim \mathcal{X} \leq \dim \mathcal{Y}$.

The generalized anchoring is a specification of a set of mappings between the state spaces, control inputs,
and parameter sets of the template and anchor. Specifically, define 
a \emph{state reduction}, $h:\mathcal{Y}\rightarrow \mathcal{X}$, that anchors the state space,
and its right-inverse, $h^\dagger:\mathcal{X}\rightarrow\mathcal{Y}$, such that $h \circ h^\dagger = id_X$.
Let $Dh$ and $Dh^\dagger$ be the Jacobians of these maps. 
Define similarly 
a \emph{control reduction}\footnote{Note that often the control input will be a subset of the cotangent
bundle over the state space, $\mathcal{U}_X \subset T^*\mathcal{X}$ and
$\mathcal{U}_Y \subset T^*\mathcal{Y}$, i.e.\ force or torque applied to one or more coordinates. In this case,
the control embedding may be related to the state space embedding, $g:=\pi_{U,Y} (Dh^\dagger)^T$, i.e.\ 
the projection down to the appropriate coordinates of the transpose of the Jacobian of $h^\dagger$.}, 
$g:\mathcal{U}_Y \rightarrow \mathcal{U}_X$, that anchors the control input,
and its right-inverse, $g^\dagger:\mathcal{U}_X\rightarrow \mathcal{U}_Y$, such that $g \circ g^\dagger = id_{\mathcal{U}_X}$.
Finally, define a \emph{parameter} or 
\emph{morphological reduction}, $\Xi:\mathcal{P}_{Y}\rightarrow \mathcal{P}_{X}$, that anchors the parameter space,
and its right-inverse, $\Xi^\dagger:\mathcal{P}_{X}\rightarrow\mathcal{P}_{Y}$, such that $\Xi \circ \Xi^\dagger = id_{\mathcal{P}_X}$.
Collectively these six maps fully define the anchoring of $Y$ in~$X$. 

An anchoring will be called \emph{exact} if,
\begin{align}
f_Y(y,u_Y,p_Y) &= Dh^\dagger \circ f_X(h(y), g(u_Y), \Xi(p_Y)),\quad
\end{align}
which implies that,
\begin{align}
f_X(x,u_X,p_X) &= Dh \circ f_Y(h^\dagger(x), g^\dagger(u_X), \Xi^\dagger(p_X))
\end{align}
(though the reverse is not necessarily true).
By contrast, an anchoring will be called \emph{approximate} if this relationship is only approximately
true (up to some desired tolerance).

Define a \emph{template controller}, $\tau_X:\mathcal{X} \times \mathcal{P}_X \rightarrow \mathcal{U}_X$, which may
be applied by assigning $u_X = \tau_X(x,p_X)$. Similarly define an \emph{anchor controller}, 
$\tau_Y:\mathcal{Y} \times \mathcal{P}_Y \rightarrow \mathcal{U}_Y$, which may be applied by assigning 
$u_Y=\tau_Y(y,p_Y)$. The template controller may be pulled back into the anchor via the choice,
\begin{align}
\tau_Y(y,p_Y) := g^\dagger \circ \tau_X(h(y),\Xi(p_Y)).
\end{align}
An anchoring will be called \emph{passive} if this is the only control authority exerted on the anchor system.
By contrast, an anchoring will be called \emph{active} if there is an additional \emph{anchoring controller}, $\bar{\tau}_Y$, exerted in
order to achieve the exact or approximate anchoring,~i.e.,
\begin{align}
u_Y = \tau_Y(y,p_Y) + \bar{\tau}_Y(y,p_Y), 
\end{align}
where $\bar{\tau}_Y$ lies in the null space of $g$.

In this paper we consider three anchor systems: one that has a passive exact anchoring, one that has a 
passive approximate anchoring, and one that has an active exact anchoring. For the passive anchors,
$\mathcal{X}=\mathcal{Y}$ and $\mathcal{U}_X=\mathcal{U}_Y$ -- therefore the maps $h, h^\dagger, g,$ and $g^\dagger$
are all identity. The active anchor, through the additional controller, $\bar{\tau}_Y$, restricts down
to the template dynamics exactly, and so these maps are similarly uninteresting. 
Therefore this paper's focus is on the remaining anchoring functions, $\Xi$ and $\Xi^\dagger$, and on the design
of the template parameters and controllers to achieve the task.

\subsection{Alternate template controller formulations}
\label{app:altcont}

For additional robustness, the template controller may use proportional-derivative (PD) feedback on the body angle (relative to the desired final position,
$\theta_{b,f}$, and velocity, $\dot{\theta}_b=0$). The controller torque takes the form,
\begin{equation}
\tau = K_p (\theta_{b,f}-\theta_b) + K_d (0-\dot{\theta}_b), \label{eq:pdcont}
\end{equation}
subject to the limits imposed by the motor model.
Given high enough gains, the torque will saturate, producing speed-limited acceleration and current-limited braking as in the switched case;
the effective switching time (when $\tau=0$) depends on the ratio of controller gains.

The ratio of gains that produces the optimal switch is found by examining
the point where the acceleration switches signs, i.e.\ when the terms of~\eqref{eq:pdcont} are equal; plugging in
the angle and velocity at the time of switch and applying the spatiotemporal transformation~\eqref{eq:spatiotemporal}
yields the ratio for the optimal value of $\tilde{\mathbf{p}}$ (see~\trsecref{trapp:pd} for this derivation).
After scaling back to physical torques the optimally-switching gain ratio is,
\begin{equation}
\frac{{K}_d}{{K}_p} \approx \frac{0.26}{ \gamma }
\end{equation}

Servoing on the internal angle produces an equivalent formulation for the PD controller.  In the dimensioned, zero angular momentum
template with initial conditions $\theta_b = \theta_r = 0$, the connection field,~\eqref{eq:templatechain} can be integrated 
to yield $\theta_b = -\xi \theta_r$.  Starting with a PD controller servoing the body angle to a desired orientation $\theta_{b,d}$,
\begin{align}
\tau &= K_p (\theta_{b,d} - \theta_b) + K_d (0 - \dot{\theta}_b) \notag \\
&= -K_p' (\theta_{r,d} - \theta_r) - K_d' (0 - \dot{\theta}_r),
\end{align}
where $K_p' := \xi K_p$, $K_d' := \xi K_d$, and the desired appendage angle $\theta_{r,d} := -\theta_{b,d}/\xi$.  The control torque on 
the appendage is opposite in sign to the body angle controller as expected.

\subsection{Dimensionless constraints for current-limited dynamics}
\label{app:cl}

If the maximum allowable torque (equivalently motor current) is limited to some factor $\beta \in (0,1)$ less than the stall torque of the
motor, $\tau_{\ell} = \beta \tau_m$, the optimal reorientation consists of three phases: 
a constant torque phase until the acceleration becomes voltage-limited, then a phase following the speed--torque
curve of the motor until the  controlled switch at $\tilde{t}_s$, followed by 
a constant braking torque phase until $\tilde{t}_h$.  
These dynamics can be integrated to produce a current-limited equivalent to $\mathcal{R}$. 
Alternatively, equivalent functions to $\tilde{g}_h$ and $\tilde{g}_\theta$ can be used to calculate $k_p$, $k_s$ and $k_t$ given $\beta$ and a parameter set $\mathbf{p}$.
We provide those equivalent relations here; their full derivations can be found in~\trsecref{tr:currentlim},
\begin{align}
\tilde{g}_h( \tilde{\omega}_m, \tilde{t}_s, \beta) :=& \tilde{t}_s + \frac{\tilde{\omega}_m^2}{\beta} \left(1-\beta\exp\left(\frac{-(\tilde{t}_s-\tilde{t}_\ell)}{\tilde{\omega}_m^2}\right)\right) \label{eq:t_h_beta} \\
 \tilde{g}_\theta(\tilde{\omega}_m,\tilde{t}_s, \beta) :=& \tilde{\omega}_m \tilde{t}_s +{\tilde{\omega}_m^3}(\beta-1) \exp\left(\frac{1-\beta}{\beta}-\frac{\tilde{t}_s}{\tilde{\omega}_m^2}\right)\notag\\
&+\frac{\beta\tilde{\omega}_m^3}{2} \left(1- \exp\left(\frac{2(1-\beta)}{\beta}-\frac{2\tilde{t}_s}{\tilde{\omega}_m^2}\right)\right).
\end{align}

\end{document}